\documentclass[lettersize,journal]{IEEEtran}
\usepackage{amsfonts}

\usepackage{array}
\usepackage[caption=false,font=normalsize,labelfont=sf,textfont=sf]{subfig}
\usepackage{textcomp}
\usepackage{stfloats}
\usepackage{url}
\usepackage{verbatim}
\usepackage{graphicx}
\usepackage{cite}

\graphicspath{{figs/}}

\usepackage{bm}

\usepackage{amsmath}
\usepackage{amssymb}

\newcommand{\mf}{\mathbf}

\newcommand{\mr}{\mathrm}

\DeclareMathOperator*{\argmax}{argmax}

\usepackage[ruled,vlined]{algorithm2e}
\usepackage{algorithmic}
\usepackage{bbding}

\usepackage{enumitem}

\usepackage{multicol}
\usepackage{multirow}
\usepackage{booktabs}

\usepackage{xcolor, array, colortbl}

\definecolor{deepred}{rgb}{0.698,0.133,0.133}
\definecolor{blue}{rgb}{0,0,1}

\definecolor{orange}{rgb}{1,0.38,0}
\definecolor{beige}{rgb}{0.639,0.58,0.502}

\definecolor{lightgray}{rgb}{.91,.91,.91}

\usepackage{tikz,xcolor,hyperref}% Make Orcid icon
\definecolor{lime}{HTML}{A6CE39}
\DeclareRobustCommand{\orcidicon}{%
    \begin{tikzpicture}
    \draw[lime, fill=lime] (0,0) 
    circle [radius=0.16] 
    node[white] {{\fontfamily{qag}\selectfont \tiny ID}};    \draw[white, fill=white] (-0.0625,0.095) 
    circle [radius=0.007];    \end{tikzpicture}
    \hspace{-2mm}}
\foreach \x in {A, ..., Z}{%
    \expandafter\xdef\csname orcid\x\endcsname{\noexpand\href{https://orcid.org/\csname orcidauthor\x\endcsname}{\noexpand\orcidicon}}
    }
\begin{document}

\title{Federated Incremental Named Entity Recognition}

\author{Duzhen Zhang$^{*}$,
        Yahan Yu$^{*}$,
        Chenxing Li$^{\dagger}$,
        Jiahua Dong, 
    and Dong Yu\orcidA{},~\IEEEmembership{Fellow,~IEEE}
            \thanks{$^{*}$ denotes equal contribution. $^{\dagger}$ denotes corresponding author.}
            \thanks{This work was supported in part by the National Nature Science Foundation of China under Grant 62133005.}
        \thanks{Duzhen Zhang is with the Institute of Automation, Chinese Academy of Sciences, Beijing, China (E-mail: bladedancer957@gmail.com). This work was done when Duzhen Zhang was interning at Tencent, AI Lab, Beijing.}
        \thanks{Chenxing Li is with the Tencent, AI Lab, Beijing, China (E-mail: chenxingli@tencent.com).}
        \thanks{Yahan Yu is with the Kyoto University, Kyoto, Japan (E-mail: yahan@nlp.ist.i.kyoto-u.ac.jp).}
        \thanks{Jiahua Dong is with the Mohamed bin Zayed University of Artificial Intelligence, Abu Dhabi, UAE (E-mail: dongjiahua1995@gmail.com).}
          \thanks{Dong Yu are with the Tencent, AI Lab, Bellevue, WA 98004 USA (E-mail: dyu@global.tencent.com).}
        }

\markboth{Journal of \LaTeX\ Class Files,~Vol.~14, No.~8, August~2021}%
{Shell \MakeLowercase{\textit{et al.}}: A Sample Article Using IEEEtran.cls for IEEE Journals}


\maketitle

\begin{abstract}
Federated Named Entity Recognition (FNER) boosts model training within each local client by aggregating the model updates of decentralized local clients, without sharing their private data. However, existing FNER methods assume fixed entity types and local clients in advance, leading to their ineffectiveness in practical applications. In a more realistic scenario, local clients receive new entity types continuously, while new local clients collecting novel data may irregularly join the global FNER training. This challenging setup, referred to here as Federated Incremental NER, renders the global model suffering from heterogeneous forgetting of old entity types from both intra-client and inter-client perspectives. To overcome these challenges, we propose a Local-Global Forgetting Defense (LGFD) model. Specifically, to address intra-client forgetting, we develop a structural knowledge distillation loss to retain the latent space's feature structure and a pseudo-label-guided inter-type contrastive loss to enhance discriminative capability over different entity types, effectively preserving previously learned knowledge within local clients. To tackle inter-client forgetting, we propose a task switching monitor that can automatically identify new entity types under privacy protection and store the latest old global model for knowledge distillation and pseudo-labeling. 
Experiments demonstrate significant improvement of our LGFD model over comparison methods.\footnote{Our code is available at \url{https://github.com/BladeDancer957/FINER}.}
\end{abstract}

\begin{IEEEkeywords}
Named Entity Recognition, Federated Learning, Incremental Learning, Heterogeneous Forgetting.
\end{IEEEkeywords}

\section{Introduction}
\IEEEPARstart{F}{ederated} Learning (FL)~\cite{mcmahan2017communication,karimireddy2020scaffold} is a privacy-preserving machine learning framework where data is locally stored, and a global server coordinates distributed local clients to collaboratively train a global model by aggregating the local model updates.
Inspired by FL, recent efforts have introduced a privacy-preserving NER method named Federated NER (FNER)~\cite{ge2020fedner,zhao2021federated}, particularly in the medical domain. Medical NER involves extracting medical entities (\emph{e.g.}, drug names, symptoms, etc.) from unstructured texts, widely used in numerous intelligent healthcare tasks~\cite{wang2013rational}.
Due to the limited labeled data available from individual medical platforms and the inability to directly share data due to the sensitive patient information they contain, acquiring sufficient data to train medical NER models is challenging. FNER addresses this issue by leveraging labeled data from various medical platforms to enhance medical NER model training on each platform without exchanging the raw data.

\begin{figure}[t]
\centering
  \includegraphics[width=1.0\linewidth]{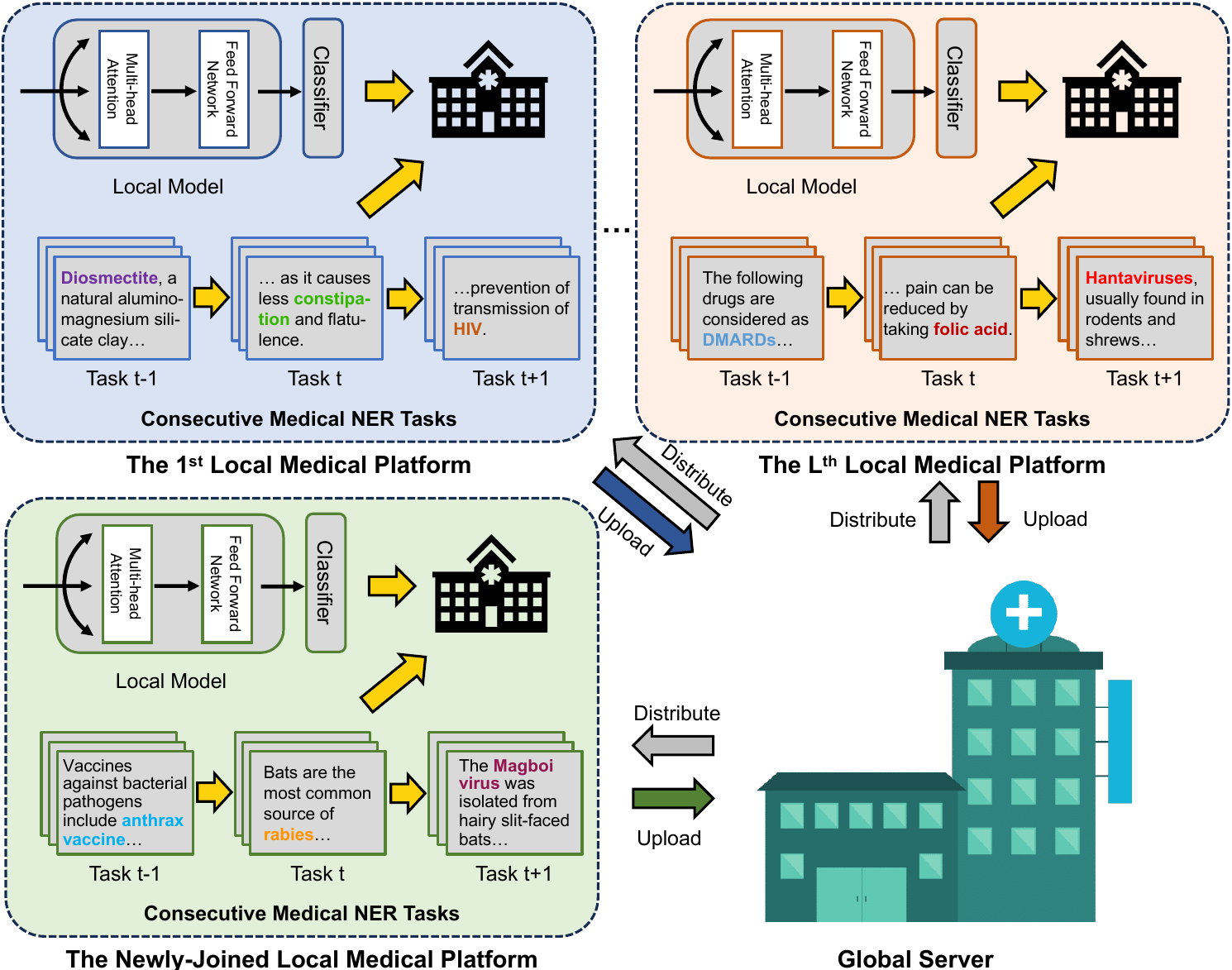} 
    \caption{Exemplary FINER setup for medical NER. Multiple medical platforms (\emph{e.g.}, hospitals) including newly-joined ones receive new entity types incrementally based on their individual preferences. FINER aims to identify novel medical entities consecutively by collaboratively learning a global medical NER model on private data of different medical platforms.}
\label{fig:intro}
\end{figure}

However, existing FNER methods unreasonably assume that entity types and local clients remain static and fixed over time, rendering them ineffective in real-world dynamic applications. In a more realistic scenario, local clients receive streaming data of new entity types consecutively, while new clients collecting novel data incrementally may irregularly join the global training of FNER.
Here, we define this challenging setup as Federated Incremental NER (FINER). Before formally defining it, we use medical NER~\cite{tang2013recognizing} as an example to better illustrate FINER, as depicted in Figure~\ref{fig:intro}.
Numerous medical platforms possess their own annotated medical NER datasets. Considering privacy protection, these platforms hope to obtain a global NER model via FL without accessing each other's private data. Over time, the data of each platform continuously expands with new medical entities appearing, and new platforms bringing novel data unseen by others join this learning process.

Formally, in the FINER setup, considering the memory overhead and privacy risks of storing old data, training data of old entity types is unavailable for all local clients. This necessitates local models to incrementally identify new entity types as they emerge, without the need for retraining on all data, thereby causing intra-client forgetting of old entity types~\cite{catastrophic_1,catastrophic_2,catastrophic_3,catastrophic_4}. This issue is further exacerbated since in each incremental learning task, old entity types from previous tasks are collapsed into the current non-entity type, leading to a phenomenon called the semantic shift of the non-entity type~\cite{zhang2023task,zhang2023continual}. Moreover, the entity type distributions are Non-Independent and Identically Distributed (Non-IID) across different local clients, causing inter-client forgetting problem. FINER aims to train a global incremental NER model via collaborative FL training on local clients while addressing heterogeneous forgetting of old entity types from both intra-client and inter-client aspects.

A naive solution to the FINER setup is to directly integrate INER~\cite{monaikul2021continual} and FL~\cite{ho2013more,yang2019federated}. However, such a simplistic approach requires the global server to possess strong prior knowledge about which local clients can collect new entity types and when, so that the global model learned in the latest old task can be stored by local clients to address forgetting on old entity types by knowledge distillation~\cite{44873_Distilling}.
Considering privacy preservation in the FINER setup, this privacy-sensitive prior knowledge cannot be shared between local clients and the global server.

To overcome the challenges mentioned above, we propose a novel Local-Global Forgetting Defense (LGFD) model to alleviate heterogeneous forgetting of old entity types from both intra-client and inter-client perspectives. Specifically, to address intra-client forgetting, we propose distilling the feature structure of the latent space from the old model to the new one by a structural knowledge distillation loss. This structure contains prominent features to effectively preserve previously learned knowledge in local clients. Additionally, we apply a pseudo-labeling strategy to identify old entity types within the current non-entity type and address the semantic shift issue. With the guidance of pseudo-labels, we design an inter-type contrastive loss to enhance discriminative capability across different entity types after pseudo-labeling and further mitigate intra-client forgetting. Furthermore, to tackle inter-client forgetting, we develop a task switching monitor to automatically identify new entity types without any prior knowledge and store the latest old model from a global perspective for knowledge distillation and pseudo-labeling. Experiments on NER datasets demonstrate significant improvement of our model over comparison methods.

Our contributions can be summarized as follows:
\begin{itemize}

\item We define a practical FINER setup and establish a corresponding benchmark. Moreover, we propose a LGFD model to address the FINER problem by overcoming heterogeneous catastrophic forgetting from both intra-client and inter-client perspectives. This represents a pioneering attempt to explore a global incremental NER model in the FL field.

\item We develop a structural knowledge distillation loss and a pseudo-label-guided inter-type contrastive loss to mitigate intra-client forgetting of old types by effectively preserving previously learned knowledge within local clients.

\item We design a task switching monitor to address inter-client forgetting by automatically recognizing new entity types under privacy protection and storing the latest old global model for knowledge distillation and pseudo-labeling.

\end{itemize}

\section{Related Work}

\subsection{FL}

FL \cite{mcmahan2017communication,fed_bayesian, fed_average, model-agnostic, fed_speedup, fedbn, hong2021federated_debiasing} is a distributed machine learning paradigm where model training occurs locally on individual devices, and only localized updates are transmitted to a global server. This method, implemented to safeguard privacy and minimize communication costs, has undergone rapid development. For instance, \cite{convergence} presents a proximal term to compel the local model to approximate the global model, ensuring convergence in FL settings. Moreover, \cite{federated_aver} suggests a weight-based mechanism for aggregating local models in global training, facilitating efficient model aggregation while preserving privacy. Additionally, \cite{layer-wise} proposes a layer-wise aggregation strategy to mitigate computation overhead during the FL process, thereby enhancing scalability and efficiency~\cite{data}.

Inspired by these FL methods, recent endeavors have incorporated FL into NER, proposing a privacy-preserving NER method known as FNER~\cite{ge2020fedner,luboshnikov2021federated,zhao2021federated,mathew2022federated,abadeer2022flightner,wang2023federated}. Moreover, \cite{dong2022federated,dong2023federated,10323204} propose a federated class-incremental learning framework by considering global and local forgetting. However, these methods mentioned above face challenges in continuously recognizing new entity types under the FINER setup.

\subsection{INER}

The conventional NER paradigm is designed to classify each token in a sequence into a fixed set of entity types (\emph{e.g.}, Person, Date, etc.) or the non-entity type \cite{ma2016end}. This approach primarily emphasizes constructing various deep learning models to enhance NER performance in an end-to-end manner \cite{li2020survey}, such as BiLSTM-CRF \cite{DBLP:conf/naacl/LampleBSKD16} and BERT-based methods \cite{kenton2019bert}. However, in more practical scenarios, new entity types emerge periodically on demand, and the NER model should be capable of incrementally recognizing these new entity types without requiring retraining from scratch \cite{zhang2023decomposing}. Consequently, recent advancements have seen the emergence of numerous methods addressing incremental learning \cite{catastrophic_4,PACKNET,GEM,NEURIPS2024Dong} in NER, referred to as INER. These approaches aim to continuously identify new entity types as they arise, eliminating the need for complete retraining \cite{monaikul2021continual,zhang2023decomposing,ma2023learning,qiu2024incremental,yu2024flexible}.

The primary challenges faced by INER include catastrophic forgetting~\cite{catastrophic_1,catastrophic_2,catastrophic_3,catastrophic_4,french1999catastrophic,kemker2018measuring} and the semantic shift of the non-entity type~\cite{zhang2023task,zhang2023continual}. To address these challenges, ExtendNER~\cite{monaikul2021continual} distills output logits from the old model to the new one. L\&R~\cite{xia2022learn} employs a two-stage learn-and-review framework, using a similar approach to ExtendNER in the learning stage and integrating synthesized samples of old entity types during the reviewing stage. Extending these concepts, CFNER~\cite{zheng2022distilling} introduces a causal framework to distill causal effects from the non-entity type. CPFD~\cite{zhang2023continual} presents a feature distillation method to retain linguistic knowledge in attention weights and a pseudo-labeling strategy to label potential old entity types within the current non-entity type, achieving State-Of-The-Art (SOTA) performance. {However, these INER methods face limitations in effectively addressing the FINER setup due to their reliance on strong prior knowledge, particularly the access to privately-sensitive information (\emph{i.e.}, when and which local clients receive new entity types).}

{To help readers better understand the advancements in this work, we clarify the distinction between our contributions and CPFD \cite{zhang2023continual}, the existing SOTA on INER. 
While CPFD addresses the core problem of INER, our work builds upon it by introducing specific enhancements for the FINER setup. The key distinctions include adapting the model to dynamic federated learning environments and incorporating techniques (\emph{i.e.}, automatically determining when and which local clients receive new entity types) to address privacy constraints.
The intra-client challenges, such as catastrophic forgetting and the semantic shift of the non-entity type, are similar to those in INER. CPFD alleviates catastrophic forgetting by using a pooled features distillation loss, but directly distilling on original feature representations may introduce noisy or redundant information. We propose distilling the geometric structures of feature representations to enhance the distillation process. Our newly designed structural knowledge distillation loss transfers the feature structure from the old model to the new one, preserving important features and better mitigating catastrophic forgetting in local clients. Additionally, we adopt a similar pseudo-labeling strategy to CPFD to address the semantic shift of the non-entity type, while introducing a pseudo-label-guided inter-type contrastive loss to improve type discrimination after pseudo-labeling, thereby further mitigating intra-client forgetting.}

\subsection{Language Language Models}

{Large Language Models (LLMs) have recently achieved significant success, demonstrating impressive performance across diverse tasks. Recent advancements emphasize scaling model size, enhancing few-shot and zero-shot capabilities, and improving task generalization. Models like GPT-4 \cite{achiam2023gpt} showcase advanced language understanding and generation, while PaLM-2 \cite{anil2023palm} optimizes multilingual processing and reasoning. Additionally, multi-modal LLMs such as GPT-4V \cite{achiam2023gpt} integrate vision and language capabilities, advancing performance on tasks requiring multi-modal comprehension and generation \cite{zhang2024mm}.}

{Given that the FINER setup we propose is model-agnostic, various language model architectures can serve as the backbone. Here, we focus on understanding-oriented language models like BERT \cite{kenton2019bert}, compared with the above generative LLMs, which align better with FINER’s specific needs for the following reasons: (1) Effectiveness in Few-Shot NER: Recent studies \cite{zhu2024llms} indicate that LLMs like GPT-4 \cite{achiam2023gpt} are less effective as zero-shot or few-shot information extractors (\emph{e.g.}, relation extraction and NER) compared to fine-tuned SOTA models. (2) Model Suitability: LLMs excel in tasks like dialogue and question answering; however, NER, typically a sequence-labeling task, is better suited to models like BERT. (3) Cost Efficiency: While LLMs can handle NER through fine-tuning, framing NER as a sequence transduction task, this approach incurs far greater training and inference costs than the BERT-based sequence-labeling method used in this paper, significantly impacting the practicality and deployment of FINER.}

\section{The Definition of the FINER Setup}

In this section, we formally define the proposed FINER setup. We begin by introducing the existing INER setup, as documented in previous works~\cite{monaikul2021continual,zheng2022distilling}.

In the INER setup, a set of incremental tasks $\mathcal{T}$=$\{\mathcal{T}^t\}_{t=1}^T$ is defined, where each task $\mathcal{T}^t$=$\{{x}_i^t, {y}_i^t\}_{i=1}^{N^t}$ comprises $N^t$ pairs of input token sequences ${x}_i^t\in\mathbb{R}^{|{x}_i^t|}$ and corresponding label sequences ${y}_i^t\in\mathbb{R}^{|{x}_i^t|}$, where $|{x}_i^t|$ represents the length of the $i$-th sequence.
The label space $\mathcal{Y}^t$ for each incremental task $t$ contains only labels corresponding to the new entity types $\mathcal{E}^t$. All other labels, such as those for potential old entity types $\mathcal{E}^{1:t-1}$ or future entity types $\mathcal{E}^{t+1:T}$, are collapsed into the non-entity type $e_{o}$. The entity types learned across different incremental tasks are non-overlapping.

We then extend the INER setup to FINER as follows: let $\mathcal{S}_g$ denote the global server and $M$ local clients be represented as $\{\mathcal{S}_m\}_{m=1}^M$. In FINER, during the $r$-th ($r$=$1,\cdots,R$) global round, we randomly select some local clients to aggregate gradients.
When the $m$-th local client is chosen to learn the $t$-th INER task, the latest global model $\Theta^{r, t}$ is distributed to $\mathcal{S}_m$ and trained on its private training data $\mathcal{T}_m^t$=$\{{x}_{mi}^{t},{y}_{mi}^t\}_{i=1}^{N_m^t}\sim\mathcal{P}_m$ of $\mathcal{S}_m$. Here, $\mf{x}_{mi}^{t}$ and $\mf{y}_{mi}^t\in\mathcal{Y}_m^t$ denote the $i$-th input token sequence and label sequence of the $m$-th local client, respectively. The distributions $\{\mathcal{P}_m\}_{m=1}^M$ are Non-IID across local clients.
The label space $\mathcal{Y}_m^t$ of $\mathcal{S}_m$ in the $t$-th task is composed of new entity types $\mathcal{E}_m^t$, where $\mathcal{Y}_m^t\subset\mathcal{Y}^t=\cup_{m=1}^M\mathcal{Y}_m^t$ and $\mathcal{E}_m^t\subset\mathcal{E}^t=\cup_{m=1}^M\mathcal{E}_m^t$). 
After obtaining the latest global model $\Theta^{r, t}$ and performing local training on $\mathcal{T}_m^t$, client $\mathcal{S}_m$ acquires an updated local model $\Theta_{m}^{r, t}$. Subsequently, the global server $\mathcal{S}_g$ aggregates the updated local models of selected clients to form the global model $\Theta^{r+1, t}$ for the training of the next global round.

In the $t$-th INER task, all local clients $\{\mathcal{S}_m\}_{m=1}^M$ are categorized into three groups: $\{\mathcal{S}_m\}_{m=1}^M = \mathbf{S}_o \cup \mathbf{S}_c \cup \mathbf{S}_n$. 
Specifically, $\mathbf{S}_o$ consists of $M_o$ local clients that have accumulated experience from previous tasks but cannot collect new training data for the $t$-th task. 
$\mathbf{S}_c$ comprises $M_c$ local clients that can receive new training data for the current task and have experience with old entity types from previous task. 
$\mathbf{S}_n$ contains $M_n$ new local clients with unseen novel entity types but without prior learning experience of old ones. These local client categories are randomly determined for each INER task.
New local clients $\mathbf{S}_n$ are added randomly during any global round in FINER, gradually increasing $M=M_o+M_c+M_n$ as incremental tasks. Importantly, there is no prior knowledge regarding the entity type distributions $\{\mathcal{P}_m\}_{m=1}^M$, the quantity and order of INER tasks, or when and which local clients will receive new entity types. In this paper, FINER aims to learn a global model $\Theta^{R, T}$ continuously to recognize new entity types while addressing heterogeneous forgetting of old entity types from both intra-client and inter-client perspectives, all while preserving the privacy of local clients.

{Finally, our approach is based on an encoder-only BERT \cite{kenton2019bert} model, which excels at capturing deep contextual relationships between input tokens and is well-suited for natural language understanding tasks (\emph{e.g.}, NER). We model NER as a sequence labeling task, specifically as token classification, and use the ``BIO" tagging scheme. In this schema, each entity type is assigned two labels: B-entity (indicating the beginning of an entity) and I-entity (indicating the continuation of an entity). For simplicity and clarity in the methods section below, we focus on entity type labels in our examples and have omitted position labels.}

\section{The Proposed LGFD Model}

\begin{figure*}[t]
	\centering
  \includegraphics[width=1.0\linewidth]{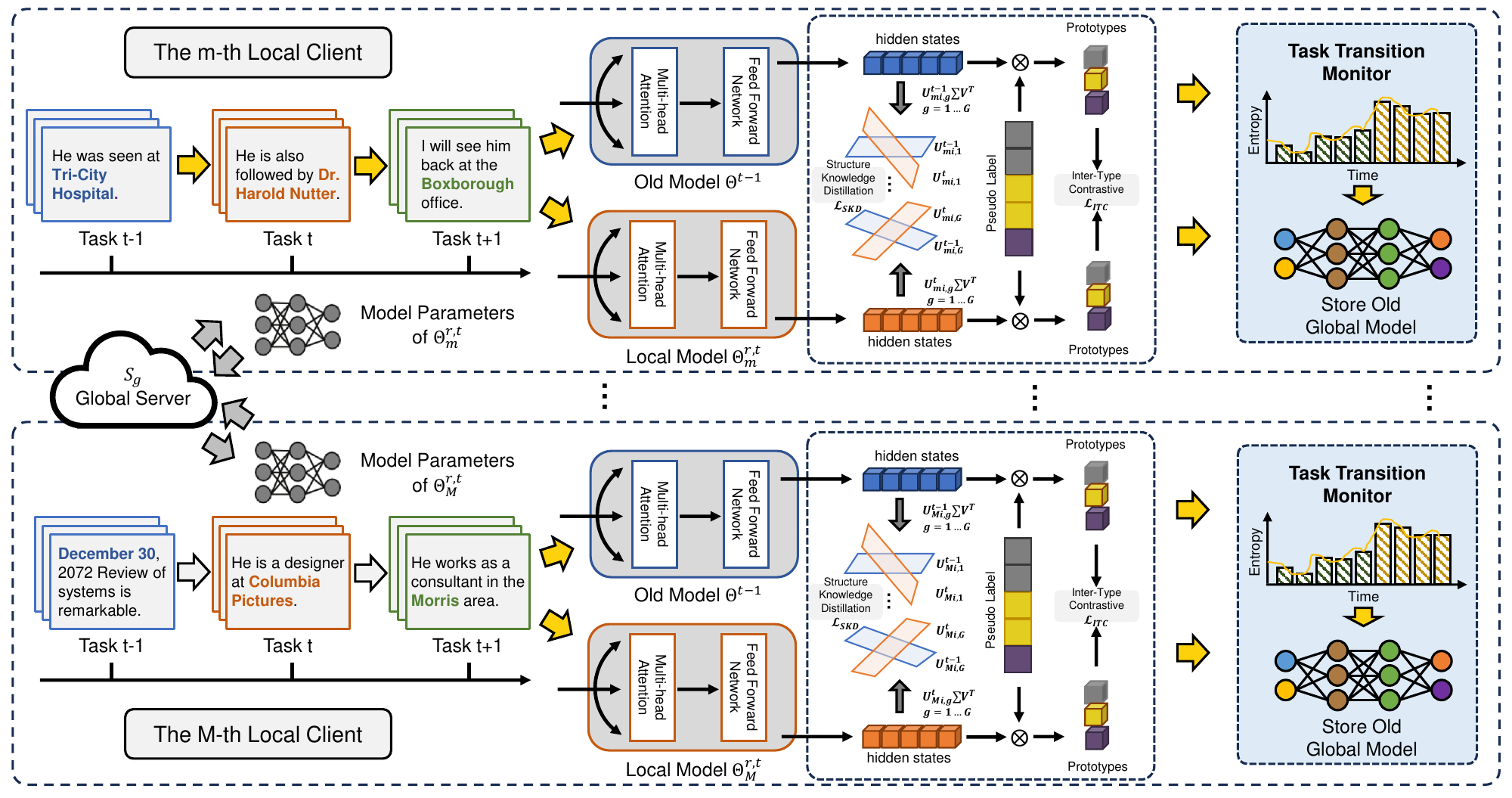}
\caption{The overview of our proposed LGFD model. It contains a \textit{structural knowledge distillation loss} $\mathcal{L}_{\mathrm{SKD}}$ and a \textit{pseudo-label-guided inter-type contrastive loss} $\mathcal{L}_{\mathrm{ITC}}$ to address intra-client forgetting by preserving previously learned knowledge within local clients. Furthermore, it employs a \textit{task switching monitor} to mitigate inter-client forgetting by automatically identifying new entity types while ensuring privacy protection and storing the latest old global model for knowledge distillation and pseudo-labeling.}
	\label{fig:model}
\end{figure*}

Figure~\ref{fig:model} depicts the overview of our proposed LGFD model designed to tackle the FINER setup. It mitigates intra-client forgetting by a Structural Knowledge Distillation (SKD) loss (Section \ref{sec:struct}) and a pseudo-label-guided Inter-Type Contrastive (ITC) loss (Section \ref{sec:inter}). Additionally, it tackles inter-client forgetting via a task switching monitor (Section \ref{sec:switch}) to automatically identify new entity types and store the latest old model from a global perspective for knowledge distillation and pseudo-labeling. The optimization pipeline of our LGFD model is detailed in Section \ref{sec:optim}.

\subsection{Structural Knowledge Distillation}\label{sec:struct}

Previous INER methods~\cite{monaikul2021continual,zheng2022distilling} commonly utilize the Knowledge Distillation (KD)~\cite{44873_Distilling} technique to mitigate the problem of forgetting old entity types. This involves distilling output logits from the old model to the new model, thereby preventing significant changes in its parameters.
Drawing inspiration from recent discoveries~\cite{clark2019does} suggesting that intermediate features, such as hidden states or attention weights, acquired by Pre-trained Language Models (PLMs) encapsulate extensive linguistic knowledge, including coreference and syntactical information, that is pivotal for NER tasks, CPFD~\cite{zhang2023continual} introduces a feature distillation loss. This loss is tailored to promote the transfer of linguistic knowledge in intermediate features, achieving SOTA INER performance.

In FINER, for the $m$-th local client $\mathcal{S}_m\in\mathbf{S}_c\cup\mathbf{S}_n$, given the $i$-th sequence pair $\{{x}_{mi}^t, {y}_{mi}^t\}$ from the $t$-th incremental task $\mathcal{T}_m^t$, we feed ${x}_{mi}^t$ into the old global model $\Theta^{t-1}$ from the last task and current local model $\Theta_{m}^{r,t}$ to obtain the hidden states of PLMs, denoted as $\mf{H}_{mi}^{t-1}$ and $\mf{H}_{mi}^{t}\in\mathbb{R}^{|x^t_{mi}|\times d_h}$, respectively.
Here, $d_h$ denotes the dimension of the hidden states and we drop the layer index for brevity.
The vanilla feature distillation loss $\mathcal{L}_{\mr{FD}}$ can be formulated as follows:
 \begin{equation}
    \mathcal{L}_{\text{FD}} = \textit{MSE}(\mf{H}^{t-1}_{mi},\ \ \mf{H}^{t}_{mi})\text{,}
      \label{fd}
 \end{equation}
where $\textit{MSE}(\cdot)$ denotes the mean squared
error loss. 

Instead of directly matching old feature representations with corresponding new ones, a process that may introduce noisy or redundant information, our hypothesis is that distilling the geometric structures of feature representations can enhance the vanilla distillation process in Equation~(\ref{fd}). This approach is expected to better aid in mitigating intra-client forgetting in the context of FINER. To achieve this goal, we introduce a SKD loss to preserve the feature structure of the latent space between models trained on incremental learning tasks. The process begins by decomposing the extracted intermediate features (\emph{e.g.}, hidden states) of PLMs and then constructing manifold structures from them by selecting prominent features. This allows us to impose constraints to maintain similar feature structures between old and new models.

Specifically, we utilize Singular Value Decomposition (SVD) to identify the principal dimensions of the feature distribution, thereby mapping the original high-dimensional features into a lower-dimensional subspace.
Formally, given original feature representations $\mf{H}_{mi}^{t-1}$ and $\mf{H}_{mi}^{t}$, we first apply SVD to obtain subspace structures $\mf{U}_{mi}^{t-1}$ and $\mf{U}_{mi}^{t}$, and then utilize them accordingly for knowledge distillation, formulated as follows:
 \begin{equation}
 \begin{aligned}
    \mathcal{L}_{\text{SKD}} &= \frac{1}{G}\sum_{g=1}^G\textit{MSE}(\mf{U}^{t-1}_{mi,g},\ \ \mf{U}^{t}_{mi,g})\\
    \mf{H}^\tau_{mi,g} &= \mf{U}^\tau_{mi,g}\mf{\Sigma}\mf{V}^T
 \end{aligned}\text{,}
      \label{skd}
 \end{equation}
where we partition the full feature representations $\mf{H}_{mi}^{\tau}$ into several smaller groups $\{\mf{H}_{mi,g}^{\tau}\}_{g=1}^G$ to reduce computational cost, $\mf{H}_{mi,g}^{\tau}\in\mathbb{R}^{|x^{t}_{mi}|\times (d_h/G)}$, $\mf{U}^\tau_{mi,g}\in\mathbb{R}^{|x^{t}_{mi}|\times |x^{t}_{mi}|}$, and $\tau$=$t$-$1$ or $t$. It's worth noting that backpropagation has been proven effective in such decomposition operations and is widely employed in locally modeling the data manifold~\cite{simon2020adaptive,roy2023subspace}.

\subsection{Contrast Inter-type Representations}\label{sec:inter}

\begin{figure}[t!]
\centering
  \includegraphics[width=1.0\linewidth]{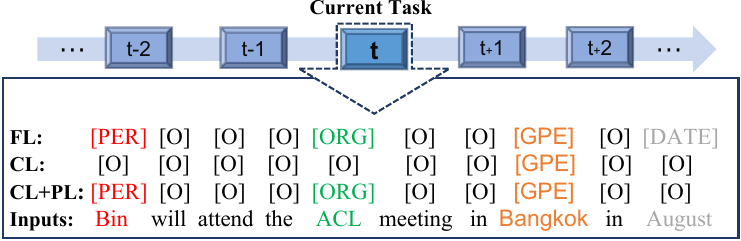} 
    \caption{Example of semantic shift of the non-entity type. \textbf{FL}, \textbf{CL}, and \textbf{PL} denote Full ground-truth Labels, Current ground-truth Labels, and Pseudo Labels. Old entity types (such as \textcolor{green}{[ORG]} (\texttt{Organization}) and \textcolor{red}{[PER]} (\texttt{Person})) and future entity types (such as \textcolor{gray}{[DATE]} (\texttt{Date})) are masked as \textcolor{black}{[O]} (\texttt{the non-entity type}) at the current step $t$ where \textcolor{orange}{[GPE]} (\texttt{Cities}) is the current entity type to be learned, leading to the semantic shift problem of the non-entity type (the second row \textbf{CL}).}
\label{fig:pseudo}
\end{figure}

Due to the limitations of data acquisition in FINER, a current ground-truth label sequence $y^t_{mi}$ only includes labels for the current entity types $\mathcal{E}^t_m$, while all other labels (\emph{e.g.}, potential old entity types $\mathcal{E}^{1:t-1}_m$ or future entity types $\mathcal{E}^{t+1:T}_m$) are collapsed into the non-entity type $e_{o}$. This leads to the semantic shift phenomenon of the non-entity type \cite{zhang2023task,zhang2023continual}. As depicted in Figure~\ref{fig:pseudo} (the second row \textbf{CL}), the old entity types \textcolor{green}{[ORG]} (\texttt{Organization}) and \textcolor{red}{[PER]} (\texttt{Person}) (learned in the prior steps such as $t$-$1$, $t$-$2$, etc.) as well as the future entity type \textcolor{gray}{[DATE]} (\texttt{Date}) (to be learned in the future steps such as $t$+$1$, $t$+$2$, etc.) are all marked as the non-entity type at the current step $t$, where \textcolor{orange}{[GPE]} (\texttt{Cities}) is the current entity type to be learned. 
Without mechanisms to distinguish tokens related to old entity types from the genuine non-entity type, this semantic shift could intensify the intra-client forgetting.

To this end, we employ a confidence-based pseudo-labeling strategy to augment the current ground-truth labels by incorporating predictions from the old model (depicted in the third row \textbf{CL+PL} in Figure \ref{fig:pseudo}), thereby generating more informative labels for classification.
Formally, we denote the cardinality of the current entity types as $E_m^t = \text{card}(\mathcal{E}_m^t)$. The predictions of the current local model $\Theta^{r,t}_m$ are represented by $\mf{\hat{y}}^t_{mi}\in\mathbb{R}^{|x^t_{mi}|\times(1+E_m^1+...+E_m^t)}$. 
We define $\mf{\tilde{y}}^t_{mi}\in\mathbb{R}^{|x^t_{mi}|\times(1+E_m^1+...+E_m^t)}$ as the target, computed using the one-hot ground-truth label sequence $\mf{y}^t_{mi}\in\mathbb{R}^{|x^t_{mi}|\times(1+E_m^1+...+E_m^t)}$ and pseudo-labels extracted from the predictions $\mf{\hat{y}}^{t-1}_{mi}\in\mathbb{R}^{|x^t_{mi}|\times(1+E^1+...+E^{t-1})}$ of the old global model $\Theta^{t-1}$ from the last task. The process is formulated as follows (omitting the subscript $mi$ for brevity):
\begin{equation}
\footnotesize
\mf{\tilde{y}}^t_{j,e}= \mkern-5mu \left\{\begin{array}{ll}
\mkern-10mu 1 \mkern-27mu & \ \ \ \text {if}\ \ \mf{y}^{t}_{j,e_{o}}=0\ \ \text {and}\ \ e = \argmax \limits_{e' \in \mathcal{E}_m^{t}}\ \mf{y}^{t}_{j,e'} \\
\mkern-10mu 1 \mkern-27mu & \ \ \ \text {if}\ \ \mf{y}^{t}_{j,e_{o}}=1\ \ \text {and}\ \ e = \mkern-8mu \argmax \limits_{e' \in e_o\cup\mathcal{E}_m^{1:t-1}} \mkern-6mu\ \ \mf{\hat{y}}^{t-1}_{j,e'}\ \text{and}\ u<\alpha_e \\
\mkern-10mu 0 \mkern-27mu & \ \ \ \text {otherwise}\\
\end{array}\right.
\label{eq:pseudo_better}
\end{equation}
In other words, if a token is not labeled as $e_{o}$, we copy the ground-truth labels (the first row in Equation (\ref{eq:pseudo_better})). Otherwise, we utilize the pseudo-labels predicted by the old model (the second row in Equation (\ref{eq:pseudo_better})). Moreover, to reduce the prediction errors from the old model, this confidence-based pseudo-labeling strategy employs entropy as a measure of uncertainty, with the median entropy serving as the confidence threshold. It retains pseudo-labels only when the old model is ``confident" enough ($u<\alpha_e$). Finally, the Cross-Entropy (CE) loss after pseudo-labeling can be expressed as follows:
\begin{equation}
    \mathcal{L}_{\text{CE}}=- \ \mf{\tilde{y}}^t_{mi}\log\mf{\hat{y}}^t_{mi}
\label{eq:pseudo_loss}
\end{equation}

As previously noted, due to the FINER setup, there is a limitation in acquiring entity type labels in the ground truth, making it difficult for the FINER model to distinguish between entity types. To enhance the model's type discrimination ability after pseudo-labeling, we design a pseudo-label-guided ITC loss. 
Given the hidden states $\mf{H}^{t-1}_{mi}$ and $\mf{H}^{t}_{mi}$, and binary masks $\tilde{\mf{y}}^t_{mi}$, we compute prototypes of each entity type as $\mf {P}_{mi}^{t,e}$ and $\mf {P}_{mi}^{t-1,e}$ (\emph{i.e.}, entity type centroids) for contrast inter-type representations (omitting the subscript $mi$ for brevity):
\begin{equation}
	{\mf{P}}^{\tau,e}=\frac{\sum\big[\mf{H}^{\tau}_{j}*\mathbb{I}\big (e=\argmax \limits_{e'\in \mathcal{E}_m^{1:t}} \mf{\tilde{y}}^{t}_{j,e'}\big)\big]}{\sum \mathbb{I}\big (e=\argmax \limits_{e'\in \mathcal{E}_m^{1:t}}\mf{\tilde{y}}^{t}_{j,e'}\big)}\text{,}
	\label{prototype}
\end{equation}
where $\tau$=$t-1$ or $t$, and $\mathbb{I}$ is the indicator function.
In the ITC loss, prototypes clustered from the same pseudo label are assigned as positive pairs in contrastive learning, while prototypes from different labels serve as negative pairs. The distance between prototypes is measured using the inner product. Consequently, the pseudo-label-guided ITC loss can be expressed as follows (omitting the subscript $mi$ for brevity):
\begin{equation}
    \mathcal{L}_{\text{ITC}}=-\frac{1}{K}\sum_{e}\log\frac{\exp(\mf{P}^{t,e}\cdot\mf{P}^{t-1,e})}{\sum_{k\neq e}\exp(\mf{P}^{t,e}\cdot\mf{P}^{*,k})},
\end{equation}
where $\mf{P}^{*} =\mf{P}^{t}\cup\mf{P}^{t-1}$, $K=|\mf{P}^t|$ is the number of inter-type prototypes.

The pseudo-label-guided ITC loss mitigates catastrophic forgetting by constructing positive contrast pairs, ensuring consistency between the corresponding prototypes extracted from hidden states $\mathbf{H}^{t-1}$ and $\mathbf{H}^{t}$. This KD-like mechanism aids in further alleviating intra-client forgetting and enhancing performance in the FINER setup.

Finally, the overall optimization objective of the selected $m$-th local client $\mathcal{S}_m$ to learn the $t$-th ($t$$\geq$$2$) INER task $\mathcal{T}_m^t$ is expressed as:
\begin{equation}
	\mathcal{L}(\Theta^{r,t}_m)= \mathcal{L}_{\mr{CE}}+\lambda_1\mathcal{L}_{\mr{SKD}} + \lambda_2 \mathcal{L}_{\mr{ITC}},
	\label{eq:overall_optimization}
\end{equation}	
where $\lambda_1, \lambda_2$ are trade-off hyper-parameters.

\subsection{Task Switching Monitor}\label{sec:switch}
When local clients incrementally recognize new entity types using Equation (\ref{eq:overall_optimization}), the global server $\mathcal{S}_g$ needs to automatically identify when and which local clients collect new entity types. Subsequently, it stores the latest old global model $\Theta^{t-1}$ for executing $\mathcal{L}_{\text{SKD}}$ and $\mathcal{L}_{\text{ITC}}$. Accurate selection of the old model $\Theta^{t-1}$ from a global perspective is essential for addressing inter-client forgetting across different local clients brought by Non-IID distributions when new entity types arrive. However, due to privacy preservation concerns, we lack prior knowledge about when to obtain new entity types in local clients under the FINER setup. To this end, we design a task switching monitor to automatically recognize when and which local clients collect new entity types. At the $r$-th round, when $\mathcal{S}_m$ receives the global model $\Theta^{r,t}$, it evaluates the average entropy $\mathcal{I}_m^{r,t}$ on $\mathcal{T}_m^t$:
\begin{equation}
	\mathcal{I}_m^{r, t} = \frac{1}{N_m^t}\sum_{i=1}^{N_m^t}\sum_{j=1}^{|x^t_{mi}|}\mathcal{H}(P_m^t({x}_{mi}^t, \Theta^{r,t})_j)\text{,}
	\label{equ:task_transition}
\end{equation}
where $\mathcal{H}(P_m^t({x}_{mi}^t,\Theta^{r,t})_j)$ represents the entropy value of the probability distribution predicted for the $j$-th token in ${x}_{mi}^t$, and $\mathcal{H}(\mf{p}) = \sum_i\mathbf{p}_i\log\mathbf{p}_i$ denotes the entropy measure function.
We consider local clients as collecting new entity types if there is a sudden increase in the averaged entropy $\mathcal{I}_m^{r,t}$, defined as $\mathcal{I}_m^{r,t} - \mathcal{I}_m^{r-1, t}\geq\lambda$, where $\lambda$ is entropy threshold and empirically set to $0.6$. Next, we update $t\leftarrow t$+$1$, and automatically store the latest global model $\Theta^{r-1, t}$ from the $(r$-$1)$-th global round as the old model $\Theta^{t-1}$. This model is then used to optimize the local model $\Theta_{m}^{r,t}$ through Equation (\ref{eq:overall_optimization}).

\begin{algorithm}[tbp]
% 	\small
	\caption{Optimization Pipeline of Our LGFD Model.}
	\label{alg: pipline_FBL}
	\LinesNumbered 
	\textbf{Input:} In the $t$-th ($t\geq2$) INER task, global server $\mathcal{S}_g$ randomly select some local clients from the client pool. Let's assume it's $w$ clients in total, denoted as $\{\mathcal{S}_{m_1}, \mathcal{S}_{m_2}, \cdots, \mathcal{S}_{m_w}\}$ with their local datasets as $\{\mathcal{T}_{m_1}^t, \mathcal{T}_{m_2}^t, \cdots, \mathcal{T}_{m_w}^t\}$ at the $r$-th global round. The global server $\mathcal{S}_g$ transmits the latest global model $\Theta^{r, t}$ to selected local clients;
 
	\textcolor{deepred}{\textbf{All Local Clients:}} \\
	\For{$\mathcal{S}_m$ in $\{\mathcal{S}_1, \mathcal{S}_2, \cdots, \mathcal{S}_M\}$}{
		Calculate averaged entropy $\mathcal{I}_m^{r, t}$ of local training data $\mathcal{T}_{m}^t$ via Equation~\eqref{equ:task_transition}; \\
	}
	
	\textcolor{deepred}{\textbf{Selected Local Clients:}} \\
	Obtain $\Theta^{r, t}$ from $\mathcal{S}_g$ as the local NER model $\Theta_m^{r, t}$;\\
	\For{$\mathcal{S}_{m}$ in $\{\mathcal{S}_{m_1}, \mathcal{S}_{m_2}, \cdots, \mathcal{S}_{m_w}\}$}{
		\emph{Task} = False\;
		\If{$\mathcal{I}_{m}^{r, t} - \mathcal{I}_{m}^{r-1, t} \geq \lambda$}{
			\emph{Task} = True\;
		}
		\If{Task = \emph{True}}{
			Store the latest global model $\Theta^{r, t}$ as old model $\Theta^{t-1}$ for local client $\mathcal{S}_m$\;
		}
		\For{$\{{x}_{mi}^t, {y}_{mi}^t\}_{i=1}^{N^t_m}$ in $\mathcal{T}_{m}^t$}{
			Update local model $\Theta_{m}^{r, t}$ via Equation~\eqref{eq:overall_optimization}; \\
		}
	}
	\textcolor{deepred}{\textbf{Global Server:}} \\
	$\mathcal{S}_g$ aggregates the updates of all local models $\Theta_m^{r, t}$ as $\Theta^{r+1, t}$ for the training of next global round. 
\end{algorithm}

\subsection{Optimization Pipeline}\label{sec:optim}

 Starting from the first INER task, all local clients utilize Equation~\eqref{equ:task_transition} to compute the average entropy $\mathcal{I}_m^{r,t}$ of local training data $\mathcal{T}_m^t$ at the beginning of each global round. Subsequently, some local clients are randomly selected by the global server $\mathcal{S}_g$ to undergo local training for each global round. Following the selection, these chosen local clients utilize the task switching monitor to accurately identify new entity types. They then automatically store the global model learned at the last global round as the old model $\Theta^{t-1}$ and optimize the local model $\Theta_{m}^{r, t}$ using Equation~\eqref{eq:overall_optimization} during the $r$-th global round. 
Finally, the updated local models $\Theta_m^{r, t}$ of selected local clients are aggregated by the global server $\mathcal{S}_g$ to form the global model $\Theta^{r+1, t}$, which is then distributed to local clients for training in the next round. 
The optimization pipeline of our LGFD model for addressing the FINER setup is outlined in {Algorithm}~\ref{alg: pipline_FBL}.

\begin{table}[t]
  \centering
  \caption{The statistics for each dataset.}
  \resizebox{1.0\linewidth}{!}{
    \begin{tabular}{ccccc}
    \toprule
        Dataset  & \multicolumn{1}{l}{\# Entity Type} & \# Sample  & \multicolumn{2}{c}{Entity Type Sequence (Alphabetical Order)} \\
    \midrule\midrule
    I2B2  & 16    & 141k   & \multicolumn{2}{c}{\begin{tabular}[1]{l}AGE, CITY, COUNTRY, DATE, DOCTOR, HOSPITAL, \\  IDNUM, MEDICALRECORD, ORGANIZATION, \\PATIENT, PHONE, PROFESSION, STATE, STREET, \\USERNAME, ZIP\end{tabular}} \\
    \midrule
    OntoNotes5 & 18    & 77k   & \multicolumn{2}{c}{\begin{tabular}[1]{l}CARDINAL, DATE, EVENT, FAC, GPE, LANGUAGE,\\ LAW, LOC, MONEY, NORP, ORDINAL, ORG,\\ PERCENT, PERSON, PRODUCT, QUANTITY, TIME,\\ WORK\_OF\_ART\end{tabular}} \\
    \bottomrule
    \end{tabular}%
    }
  \label{tab:dataset_statistics}%
\end{table}%

\begin{table*}[t]
\centering
\caption{Comparison results in terms of Mi-F1 and Ma-F1 (\%) on the I2B2 dataset under the setting of $8$-$1$. The \textcolor{beige}{\textbf{beige}} denotes the old entity types that have already been learned, and the \textcolor{orange}{\textbf{orange}} denotes the new entity types that are currently being learned. The \textcolor{deepred}{\textbf{red}} represents the highest result, and the \textcolor{blue}{\textbf{blue}} represents the second highest result. The markers $\dagger$ refers to significant tests comparing with CPFD+FL ($p$-$value<0.05$).}
\resizebox{0.96\linewidth}{!}{
\begin{tabular}{c|cccc|cccc|cccc|cccc|>{\columncolor{lightgray}}c>{\columncolor{lightgray}}c}
		\toprule
		Task ID   & \multicolumn{4}{c|}{t=2}&\multicolumn{4}{c|}{t=3} &\multicolumn{4}{c|}{t=4}&\multicolumn{4}{c|}{t=5}& & \\
		Entity Type ID &\multicolumn{1}{c|}{\textcolor{beige}{\textbf{1-8}}} & \multicolumn{1}{c|}{\textcolor{orange}{\textbf{9}}}&\multicolumn{2}{c|}{All(1-9)} & \multicolumn{1}{c|}{\textcolor{beige}{\textbf{1-9}}} & \multicolumn{1}{c|}{\textcolor{orange}{\textbf{10}}}&\multicolumn{2}{c|}{All(1-10)}& \multicolumn{1}{c|}{\textcolor{beige}{\textbf{1-10}}} & \multicolumn{1}{c|}{\textcolor{orange}{\textbf{11}}}&\multicolumn{2}{c|}{All(1-11)}&\multicolumn{1}{c|}{\textcolor{beige}{\textbf{1-11}}} & \multicolumn{1}{c|}{\textcolor{orange}{\textbf{12}}}&\multicolumn{2}{c|}{All(1-12)}&Avg.& Avg.  \\
        Metrics   & \multicolumn{1}{c|}{Ma-F1}  & \multicolumn{1}{c|}{Ma-F1} & \multicolumn{1}{c}{Mi-F1}& \multicolumn{1}{c|}{Ma-F1}  & \multicolumn{1}{c|}{Ma-F1} & \multicolumn{1}{c|}{Ma-F1}& \multicolumn{1}{c}{Mi-F1}& \multicolumn{1}{c|}{Ma-F1}  &
        \multicolumn{1}{c|}{Ma-F1} & \multicolumn{1}{c|}{Ma-F1}&\multicolumn{1}{c}{Mi-F1}& \multicolumn{1}{c|}{Ma-F1}  & \multicolumn{1}{c|}{Ma-F1} & \multicolumn{1}{c|}{Ma-F1}&\multicolumn{1}{c}{Mi-F1}& \multicolumn{1}{c|}{Ma-F1} &
       Mi-F1& Ma-F1
        \\
 \hline
         FT+FL & \multicolumn{1}{c|}{28.76} & \multicolumn{1}{c|}{0.00} & 24.79 & 25.56 & \multicolumn{1}{c|}{0.45} & \multicolumn{1}{c|}{18.57} & 6.97 & 2.26 & \multicolumn{1}{c|}{0.88} & \multicolumn{1}{c|}{11.84} & 2.87 & 1.87 & \multicolumn{1}{c|}{2.61} & \multicolumn{1}{c|}{17.65} & 2.64 & 3.86 & -- & -- \\

         PODNet+FL  & \multicolumn{1}{c|}{17.21} & \multicolumn{1}{c|}{1.38} & 17.46 & 15.45 & \multicolumn{1}{c|}{2.47} & \multicolumn{1}{c|}{8.30} & 6.32 & 3.05 & \multicolumn{1}{c|}{1.51} & \multicolumn{1}{c|}{4.38} & 5.93 & 1.77 & \multicolumn{1}{c|}{1.70} & \multicolumn{1}{c|}{4.06} & 3.87 & 1.89 & -- & --\\

     LUCIR+FL & \multicolumn{1}{c|}{33.15} & \multicolumn{1}{c|}{\textcolor{blue}{\textbf{5.70}}} & \textcolor{blue}{\textbf{58.47}} & 30.10 & \multicolumn{1}{c|}{14.09} & \multicolumn{1}{c|}{21.75} & 32.25 & 14.86 & \multicolumn{1}{c|}{5.25} & \multicolumn{1}{c|}{6.71} & 8.19 & 5.38 & \multicolumn{1}{c|}{4.58} & \multicolumn{1}{c|}{21.27} & 8.16 & 5.97 & -- & -- \\

      ST+FL& \multicolumn{1}{c|}{16.45} & \multicolumn{1}{c|}{4.87} & 22.64 & 15.16 & \multicolumn{1}{c|}{3.17} & \multicolumn{1}{c|}{20.30} & 16.46 & 4.88 & \multicolumn{1}{c|}{5.79} & \multicolumn{1}{c|}{12.72} & 14.18 & 6.42 & \multicolumn{1}{c|}{5.62} & \multicolumn{1}{c|}{\textcolor{deepred}{\textbf{40.85}}} & 14.31 & 8.56 & -- & --\\

         ExtendNER+FL & \multicolumn{1}{c|}{14.17} & \multicolumn{1}{c|}{5.23} & 19.11 & 13.18 & \multicolumn{1}{c|}{2.20} & \multicolumn{1}{c|}{19.78} & 12.27 & 3.95 & \multicolumn{1}{c|}{4.13} & \multicolumn{1}{c|}{12.42} & 13.16 & 4.89 & \multicolumn{1}{c|}{4.77} & \multicolumn{1}{c|}{\textcolor{blue}{\textbf{38.55}}} & 13.99 & 7.58 & -- & -- \\

   CFNER+FL & \multicolumn{1}{c|}{34.64} & \multicolumn{1}{c|}{\textcolor{deepred}{\textbf{6.52}}} & 47.79 & 31.52 & \multicolumn{1}{c|}{14.75} & \multicolumn{1}{c|}{\textcolor{deepred}{\textbf{28.47}}} & 27.14 & 16.12 & \multicolumn{1}{c|}{9.15} & \multicolumn{1}{c|}{16.05} & 17.64 & 9.78 & \multicolumn{1}{c|}{7.31} & \multicolumn{1}{c|}{30.29} & 14.88 & 9.23 & -- & -- \\

      CPFD+FL & \multicolumn{1}{c|}{\textcolor{blue}{\textbf{39.59}}} & \multicolumn{1}{c|}{1.52} & 40.70 & \textcolor{blue}{\textbf{35.36}} & \multicolumn{1}{c|}{\textcolor{blue}{\textbf{27.16}}} & \multicolumn{1}{c|}{20.81}& \textcolor{blue}{\textbf{34.26}} & \textcolor{blue}{\textbf{26.53}} & \multicolumn{1}{c|}{\textcolor{blue}{\textbf{26.86}}} & \multicolumn{1}{c|}{\textcolor{blue}{\textbf{25.16}}} & \textcolor{blue}{\textbf{34.69}}& \textcolor{blue}{\textbf{26.71}}& \multicolumn{1}{c|}{\textcolor{blue}{\textbf{26.05}}} & \multicolumn{1}{c|}{29.06}&\textcolor{blue}{\textbf{34.66}}&\textcolor{blue}{\textbf{26.30}}& --& --\\

 \hline

 \textbf{LGFD (Ours)}   & \multicolumn{1}{c|}{\textcolor{deepred}{\textbf{51.69}}} & \multicolumn{1}{c|}{3.37} & \textcolor{deepred}{\textbf{60.26}} & \textcolor{deepred}{\textbf{46.32}} & \multicolumn{1}{c|}{\textcolor{deepred}{\textbf{34.27}}} & \multicolumn{1}{c|}{\textcolor{blue}{\textbf{22.28}}}& \textcolor{deepred}{\textbf{44.65}} & \textcolor{deepred}{\textbf{33.08}} & \multicolumn{1}{c|}{\textcolor{deepred}{\textbf{30.84}}} & \multicolumn{1}{c|}{\textcolor{deepred}{\textbf{29.12}}} & \textcolor{deepred}{\textbf{43.43}} & \textcolor{deepred}{\textbf{30.68}} & \multicolumn{1}{c|}{\textcolor{deepred}{\textbf{27.96}}} & \multicolumn{1}{c|}{36.79}&\textcolor{deepred}{\textbf{41.12}}&\textcolor{deepred}{\textbf{28.70}}& --& --\\

\hline\hline
Task ID   & \multicolumn{4}{c|}{t=6}&\multicolumn{4}{c|}{t=7} &\multicolumn{4}{c|}{t=8}&\multicolumn{4}{c|}{t=9}& & \\
		Entity Type ID &\multicolumn{1}{c|}{\textcolor{beige}{\textbf{1-12}}} & \multicolumn{1}{c|}{\textcolor{orange}{\textbf{13}}}&\multicolumn{2}{c|}{All(1-13)} & \multicolumn{1}{c|}{\textcolor{beige}{\textbf{1-13}}} & \multicolumn{1}{c|}{\textcolor{orange}{\textbf{14}}}&\multicolumn{2}{c|}{All(1-14)}& \multicolumn{1}{c|}{\textcolor{beige}{\textbf{1-14}}} & \multicolumn{1}{c|}{\textcolor{orange}{\textbf{15}}}&\multicolumn{2}{c|}{All(1-15)}&\multicolumn{1}{c|}{\textcolor{beige}{\textbf{1-15}}} & \multicolumn{1}{c|}{\textcolor{orange}{\textbf{16}}}&\multicolumn{2}{c|}{All(1-16)}&Avg.& Avg.  \\
        Metrics   & \multicolumn{1}{c|}{Ma-F1}  & \multicolumn{1}{c|}{Ma-F1} & \multicolumn{1}{c}{Mi-F1}& \multicolumn{1}{c|}{Ma-F1}  & \multicolumn{1}{c|}{Ma-F1} & \multicolumn{1}{c|}{Ma-F1}& \multicolumn{1}{c}{Mi-F1}& \multicolumn{1}{c|}{Ma-F1}  &
        \multicolumn{1}{c|}{Ma-F1} & \multicolumn{1}{c|}{Ma-F1}&\multicolumn{1}{c}{Mi-F1}& \multicolumn{1}{c|}{Ma-F1}  & \multicolumn{1}{c|}{Ma-F1} & \multicolumn{1}{c|}{Ma-F1}&\multicolumn{1}{c}{Mi-F1}& \multicolumn{1}{c|}{Ma-F1} &
       Mi-F1& Ma-F1
        \\
    \hline
         FT+FL  & \multicolumn{1}{c|}{0.46} & \multicolumn{1}{c|}{14.15} & 0.92 & 1.51 & \multicolumn{1}{c|}{0.14} & \multicolumn{1}{c|}{7.82} & 1.09 & 0.69 & \multicolumn{1}{c|}{1.25} & \multicolumn{1}{c|}{10.00} & 1.25 & 1.84 & \multicolumn{1}{c|}{0.41} & \multicolumn{1}{c|}{\textcolor{blue}{\textbf{24.23}}} & 1.46 & 1.90 & 5.25 & 4.94 \\

         PODNet+FL  & \multicolumn{1}{c|}{2.63} & \multicolumn{1}{c|}{4.52} & 3.57 & 2.77 & \multicolumn{1}{c|}{2.72} & \multicolumn{1}{c|}{0.85} & 1.09 & 2.58 & \multicolumn{1}{c|}{2.16} & \multicolumn{1}{c|}{0.83} & 1.09 & 2.07 & \multicolumn{1}{c|}{2.08} & \multicolumn{1}{c|}{5.15} & 1.61 & 2.27 & 5.12 & 3.98 \\

     LUCIR+FL & \multicolumn{1}{c|}{5.90} & \multicolumn{1}{c|}{21.07} & 8.31 & 7.06 & \multicolumn{1}{c|}{6.12} & \multicolumn{1}{c|}{5.69} & 5.86 & 6.09 & \multicolumn{1}{c|}{5.79} & \multicolumn{1}{c|}{4.85} & 5.24 & 5.73 & \multicolumn{1}{c|}{5.04} & \multicolumn{1}{c|}{17.33} & 4.37 & 5.81 & 16.36 & 10.12 \\

      ST+FL& \multicolumn{1}{c|}{7.63} & \multicolumn{1}{c|}{\textcolor{blue}{\textbf{28.17}}} & 12.23 & 9.21 & \multicolumn{1}{c|}{7.71} & \multicolumn{1}{c|}{10.73} & 10.99 & 7.93 & \multicolumn{1}{c|}{6.63} & \multicolumn{1}{c|}{\textcolor{blue}{\textbf{13.75}}} & 9.72 & 7.10 & \multicolumn{1}{c|}{7.28} & \multicolumn{1}{c|}{23.84} & 9.97 & 8.32 & 13.81 & 8.45\\

         ExtendNER+FL  & \multicolumn{1}{c|}{7.13} & \multicolumn{1}{c|}{26.69} & 11.53 & 8.64 & \multicolumn{1}{c|}{8.39} & \multicolumn{1}{c|}{\textcolor{deepred}{\textbf{10.99}}} & 10.60 & 8.58 & \multicolumn{1}{c|}{8.01} & \multicolumn{1}{c|}{13.15} & 8.54 & 8.35 & \multicolumn{1}{c|}{7.79} & \multicolumn{1}{c|}{23.03} & 9.05 & 8.74 & 12.28 & 7.99 \\

   CFNER+FL & \multicolumn{1}{c|}{8.51} & \multicolumn{1}{c|}{\textcolor{deepred}{\textbf{30.80}}} & 14.93 & 10.23 & \multicolumn{1}{c|}{8.87} & \multicolumn{1}{c|}{\textcolor{blue}{\textbf{10.92}}} & 11.01 & 9.02 & \multicolumn{1}{c|}{8.26} & \multicolumn{1}{c|}{\textcolor{deepred}{\textbf{14.97}}} & 9.13 & 8.71 & \multicolumn{1}{c|}{9.93} & \multicolumn{1}{c|}{23.36} & 10.05 & 10.77 & 19.07 & 13.17\\

      CPFD+FL &\multicolumn{1}{c|}{\textcolor{blue}{\textbf{25.64}}}& \multicolumn{1}{c|}{14.86}&\textcolor{blue}{\textbf{33.52}}&\textcolor{blue}{\textbf{24.81}}& \multicolumn{1}{c|}{\textcolor{blue}{\textbf{24.36}}}& \multicolumn{1}{c|}{3.04}&\textcolor{blue}{\textbf{32.95}}&\textcolor{blue}{\textbf{22.84}}&\multicolumn{1}{c|}{\textcolor{blue}{\textbf{25.03}}}&\multicolumn{1}{c|}{5.71}&\textcolor{blue}{\textbf{34.31}}&\textcolor{blue}{\textbf{23.74}}&\multicolumn{1}{c|}{\textcolor{blue}{\textbf{23.22}}}&\multicolumn{1}{c|}{\textcolor{deepred}{\textbf{38.20}}}&\textcolor{blue}{\textbf{33.35}}&\textcolor{deepred}{\textbf{24.16}}&\textcolor{blue}{\textbf{34.80}}&\textcolor{blue}{\textbf{26.30}} \\

 \hline

     \textbf{LGFD (Ours)}  &\multicolumn{1}{c|}{\textcolor{deepred}{\textbf{28.45}}}& \multicolumn{1}{c|}{16.93}&\textcolor{deepred}{\textbf{40.76}}&\textcolor{deepred}{\textbf{27.56}}& \multicolumn{1}{c|}{\textcolor{deepred}{\textbf{27.56}}}& \multicolumn{1}{c|}{3.90}&\textcolor{deepred}{\textbf{40.70}}&\textcolor{deepred}{\textbf{25.87}}&\multicolumn{1}{c|}{\textcolor{deepred}{\textbf{25.34}}}&\multicolumn{1}{c|}{5.54}&\textcolor{deepred}{\textbf{40.00}}&\textcolor{deepred}{\textbf{24.02}}&\multicolumn{1}{c|}{\textcolor{deepred}{\textbf{23.25}}}&\multicolumn{1}{c|}{4.05}&\textcolor{deepred}{\textbf{39.39}}&\textcolor{blue}{\textbf{22.05}}&$\textbf{\textcolor{deepred}{43.79}}^{\dagger}$&$\textbf{\textcolor{deepred}{29.79}}^{\dagger}$ \\

\hline

        \textbf{Upper Bound}  &\multicolumn{1}{c|}{--}& \multicolumn{1}{c|}{--}&--&--& \multicolumn{1}{c|}{--}& \multicolumn{1}{c|}{--}&--&--&\multicolumn{1}{c|}{--}&\multicolumn{1}{c|}{--}&--&--&\multicolumn{1}{c|}{--}&\multicolumn{1}{c|}{--}&88.99&79.12&--&-- \\

\bottomrule

\end{tabular}
}

\label{tab:i2b2_8-1}
\end{table*}

\begin{table*}[t]
\centering
\caption{Comparison results in terms of Mi-F1 and Ma-F1 (\%) on the I2B2 dataset under the setting of $8$-$2$. The \textcolor{beige}{\textbf{beige}} denotes the old entity types that have already been learned, and the \textcolor{orange}{\textbf{orange}} denotes the new entity types that are currently being learned. The \textcolor{deepred}{\textbf{red}} represents the highest result, and the \textcolor{blue}{\textbf{blue}} represents the second highest result. The markers $\dagger$ and $\ddagger$ refer to significant tests comparing with CPFD+FL and CFNER+FL, respectively ($p$-$value<0.05$).}
\resizebox{0.96\linewidth}{!}{
\begin{tabular}{c|cccc|cccc|cccc|cccc|>{\columncolor{lightgray}}c>{\columncolor{lightgray}}c}
		\toprule
		Task ID   & \multicolumn{4}{c|}{t=2}&\multicolumn{4}{c|}{t=3} &\multicolumn{4}{c|}{t=4}&\multicolumn{4}{c|}{t=5}& & \\
		Entity Type ID &\multicolumn{1}{c|}{\textcolor{beige}{\textbf{1-8}}} & \multicolumn{1}{c|}{\textcolor{orange}{\textbf{9-10}}}&\multicolumn{2}{c|}{All(1-10)} & \multicolumn{1}{c|}{\textcolor{beige}{\textbf{1-10}}} & \multicolumn{1}{c|}{\textcolor{orange}{\textbf{11-12}}}&\multicolumn{2}{c|}{All(1-12)}& \multicolumn{1}{c|}{\textcolor{beige}{\textbf{1-12}}} & \multicolumn{1}{c|}{\textcolor{orange}{\textbf{13-14}}}&\multicolumn{2}{c|}{All(1-14)}&\multicolumn{1}{c|}{\textcolor{beige}{\textbf{1-14}}} & \multicolumn{1}{c|}{\textcolor{orange}{\textbf{15-16}}}&\multicolumn{2}{c|}{All(1-16)}&Avg.& Avg.  \\
        Metrics   & \multicolumn{1}{c|}{Ma-F1}  & \multicolumn{1}{c|}{Ma-F1} & \multicolumn{1}{c}{Mi-F1}& \multicolumn{1}{c|}{Ma-F1}  & \multicolumn{1}{c|}{Ma-F1} & \multicolumn{1}{c|}{Ma-F1}& \multicolumn{1}{c}{Mi-F1}& \multicolumn{1}{c|}{Ma-F1}  &
        \multicolumn{1}{c|}{Ma-F1} & \multicolumn{1}{c|}{Ma-F1}&\multicolumn{1}{c}{Mi-F1}& \multicolumn{1}{c|}{Ma-F1}  & \multicolumn{1}{c|}{Ma-F1} & \multicolumn{1}{c|}{Ma-F1}&\multicolumn{1}{c}{Mi-F1}& \multicolumn{1}{c|}{Ma-F1} &
       Mi-F1& Ma-F1
        \\
 \hline
         FT+FL  & \multicolumn{1}{c|}{4.34} & \multicolumn{1}{c|}{14.14} & 8.35 & 6.30 & \multicolumn{1}{c|}{4.09} & \multicolumn{1}{c|}{31.74} & 6.73 & 8.70 & \multicolumn{1}{c|}{1.96} & \multicolumn{1}{c|}{18.46} & 2.42 & 4.32 & \multicolumn{1}{c|}{2.88} & \multicolumn{1}{c|}{19.33} & 2.59 & 4.94 & 5.02 & 6.06 \\

         PODNet+FL  & \multicolumn{1}{c|}{17.87} & \multicolumn{1}{c|}{10.70} & 30.74 & 16.43 & \multicolumn{1}{c|}{3.05} & \multicolumn{1}{c|}{9.73} & 6.69 & 4.16 & \multicolumn{1}{c|}{4.01} & \multicolumn{1}{c|}{11.01} & 5.41 & 5.01 & \multicolumn{1}{c|}{6.06} & \multicolumn{1}{c|}{4.97} & 4.78 & 5.92 & 11.90 & 7.88\\

     LUCIR+FL & \multicolumn{1}{c|}{30.71} & \multicolumn{1}{c|}{22.11} & 55.91 & 28.99 & \multicolumn{1}{c|}{16.47} & \multicolumn{1}{c|}{25.51} & 37.82 & 17.98 & \multicolumn{1}{c|}{14.13} & \multicolumn{1}{c|}{25.88} & 28.66 & 15.81 & \multicolumn{1}{c|}{10.29} & \multicolumn{1}{c|}{13.67} & 11.22 & 10.71 & 33.40 & 18.37 \\

      ST+FL& \multicolumn{1}{c|}{22.26} & \multicolumn{1}{c|}{18.64} & 47.25 & 21.53 & \multicolumn{1}{c|}{9.81} & \multicolumn{1}{c|}{31.76} & 29.54 & 13.47 & \multicolumn{1}{c|}{11.40} & \multicolumn{1}{c|}{29.35} & 24.70 & 13.96 & \multicolumn{1}{c|}{13.18} & \multicolumn{1}{c|}{24.70} & 27.17 & 14.62 & 32.16 & 15.89\\

         ExtendNER+FL  & \multicolumn{1}{c|}{26.01} & \multicolumn{1}{c|}{19.61} & 49.36 & 24.73 & \multicolumn{1}{c|}{10.60} & \multicolumn{1}{c|}{\textcolor{blue}{\textbf{32.44}}} & 30.19 & 14.24 & \multicolumn{1}{c|}{11.73} & \multicolumn{1}{c|}{\textcolor{blue}{\textbf{30.18}}} & 25.44 & 14.37 & \multicolumn{1}{c|}{13.69} & \multicolumn{1}{c|}{\textcolor{deepred}{\textbf{30.30}}} & 26.55 & 15.77 & 32.89 & 17.28 \\

   CFNER+FL & \multicolumn{1}{c|}{\textcolor{blue}{\textbf{52.15}}} & \multicolumn{1}{c|}{\textcolor{deepred}{\textbf{25.16}}} & \textcolor{blue}{\textbf{62.15}} & \textcolor{blue}{\textbf{46.75}} & \multicolumn{1}{c|}{\textcolor{blue}{\textbf{29.40}}} & \multicolumn{1}{c|}{\textcolor{deepred}{\textbf{34.36}}} & \textcolor{blue}{\textbf{46.73}} & \textcolor{blue}{\textbf{30.22}} & \multicolumn{1}{c|}{20.65} & \multicolumn{1}{c|}{\textcolor{deepred}{\textbf{32.38}}} & 33.58 & 22.32 & \multicolumn{1}{c|}{20.87} & \multicolumn{1}{c|}{\textcolor{blue}{\textbf{28.10}}} & 30.04 & 21.77 & 43.13 & \textcolor{blue}{\textbf{30.27}} \\

      CPFD+FL & \multicolumn{1}{c|}{43.79} & \multicolumn{1}{c|}{15.71} & 51.72 & 38.18 & \multicolumn{1}{c|}{28.93} & \multicolumn{1}{c|}{26.92} & 44.48 & 28.60 & \multicolumn{1}{c|}{\textcolor{blue}{\textbf{25.38}}} & \multicolumn{1}{c|}{15.47} & \textcolor{deepred}{\textbf{41.92}} & \textcolor{blue}{\textbf{23.97}} & \multicolumn{1}{c|}{\textcolor{blue}{\textbf{22.91}}} & \multicolumn{1}{c|}{24.12} &\textcolor{deepred}{\textbf{41.12}} & \textcolor{blue}{\textbf{23.06}} & \textcolor{blue}{\textbf{44.81}} & 28.45 \\

 \hline

 \textbf{LGFD (Ours)}   & \multicolumn{1}{c|}{\textcolor{deepred}{\textbf{58.59}}} & \multicolumn{1}{c|}{\textcolor{blue}{\textbf{23.81}}} & \textcolor{deepred}{\textbf{66.53}} & \textcolor{deepred}{\textbf{51.63}} & \multicolumn{1}{c|}{\textcolor{deepred}{\textbf{33.49}}} & \multicolumn{1}{c|}{27.13} & \textcolor{deepred}{\textbf{48.08}} & \textcolor{deepred}{\textbf{32.43}} & \multicolumn{1}{c|}{\textcolor{deepred}{\textbf{26.15}}} & \multicolumn{1}{c|}{15.33} & \textcolor{blue}{\textbf{39.66}} & \textcolor{deepred}{\textbf{24.60}} & \multicolumn{1}{c|}{\textcolor{deepred}{\textbf{23.78}}} & \multicolumn{1}{c|}{21.74} & \textcolor{blue}{\textbf{38.04}} & \textcolor{deepred}{\textbf{23.53}} & $\textbf{\textcolor{deepred}{48.08}}^{\dagger}$ & $\textbf{\textcolor{deepred}{33.05}}^{\ddagger}$ \\

\hline
    \textbf{Upper Bound}  &\multicolumn{1}{c|}{--}& \multicolumn{1}{c|}{--}&--&--& \multicolumn{1}{c|}{--}& \multicolumn{1}{c|}{--}&--&--&\multicolumn{1}{c|}{--}&\multicolumn{1}{c|}{--}&--&--&\multicolumn{1}{c|}{--}&\multicolumn{1}{c|}{--}&88.99&79.12&--&-- \\

		\bottomrule
\end{tabular}
}
\label{tab:i2b2_8-2}
\end{table*}

\section{Experimental Configurations}

\subsection{Datasets} 

Following SOTA INER method~\cite{zhang2023continual}, we employ two extensively used NER datasets: I2B2 \cite{murphy2010serving} and OntoNotes5 \cite{hovy2006ontonotes} to assess the effectiveness of our LGFD model, where I2B2 is a dataset in the medical domain. Detailed statistics for these datasets are provided in Table~\ref{tab:dataset_statistics}.
We apply the same greedy sampling algorithm as suggested in CFNER~\cite{zheng2022distilling} to partition the training set into disjoint slices, each corresponding to different incremental learning tasks. Within each slice, we exclusively retain the labels associated with the entity types being learned, while masking others as the non-entity type. For a thorough explanation and detailed breakdown of this greedy sampling algorithm, please refer to Appendix B of CFNER~\cite{zheng2022distilling}.

\subsection{FINER Settings} 

For training, we introduce entity types in alphabetical order and continually train models using corresponding data slices. 
To get closer to real-world situations, we train the base model with half of all entity types, allowing the model to acquire sufficient knowledge before incremental learning.
Specifically, on the I2B2 dataset, settings $8$-$1$ and $8$-$2$ involve learning $8$ entity types followed by $1$ entity type $8$ times ($T$=$9$), and by $2$ entity types $4$ times ($T$=$5$), respectively. Similarly, on the OntoNotes5 dataset, settings $10$-$1$ and $10$-$2$ require learning $10$ entity types followed by $1$ entity type $8$ times ($T$=$9$), and by $2$ entity types $4$ times ($T$=$5$).
We initialize local clients as $10$ and add $4$ new local clients for each incremental task. For local training, we randomly select $40\%$ of local clients to undergo $5$ epochs if each incremental task ($t\geq2$) involves learning an entity type; otherwise, if the incremental task involves learning more than one entity type, they undergo $10$ epochs. 
We randomly choose $60\%$ of samples for each selected local client in each task. For testing, we retain labels for all previously learned entity types, designating the rest as the non-entity type within the test set.

\subsection{Evaluation Metrics}
Following the SOTA INER method~\cite{zhang2023continual}, we employ Micro F1 (Mi-F1) and Macro F1 (Ma-F1) scores to assess the model's performance. We provide separate Ma-F1 scores for old and new entity types in each incremental task, as well as Mi-F1 and Ma-F1 scores for all entity types within each task, along with the final average Mi-F1 and Ma-F1 scores across all tasks (\textbf{except for the first task}). To evaluate the statistical significance of the improvements, we conduct a paired t-test with a significance level of $0.05$~\cite{koehn2004statistical}.

\begin{table*}[t]
\centering
\caption{Comparison results in terms of Mi-F1 and Ma-F1 (\%) on the OntoNotes5 dataset under the setting of $10$-$1$. The \textcolor{beige}{\textbf{beige}} denotes the old entity types that have already been learned, and the \textcolor{orange}{\textbf{orange}} denotes the new entity types that are currently being learned. The \textcolor{deepred}{\textbf{red}} represents the highest result, and the \textcolor{blue}{\textbf{blue}} represents the second highest result. The markers $\dagger$ refers to significant tests comparing with CPFD+FL ($p$-$value<0.05$).}
\resizebox{1.0\linewidth}{!}{
\begin{tabular}{c|cccc|cccc|cccc|cccc|>{\columncolor{lightgray}}c>{\columncolor{lightgray}}c}
		\toprule
		Task ID   & \multicolumn{4}{c|}{t=2}&\multicolumn{4}{c|}{t=3} &\multicolumn{4}{c|}{t=4}&\multicolumn{4}{c|}{t=5}& & \\
		Entity Type ID &\multicolumn{1}{c|}{\textcolor{beige}{\textbf{1-10}}} & \multicolumn{1}{c|}{\textcolor{orange}{\textbf{11}}}&\multicolumn{2}{c|}{All(1-11)} & \multicolumn{1}{c|}{\textcolor{beige}{\textbf{1-11}}} & \multicolumn{1}{c|}{\textcolor{orange}{\textbf{12}}}&\multicolumn{2}{c|}{All(1-12)}& \multicolumn{1}{c|}{\textcolor{beige}{\textbf{1-12}}} & \multicolumn{1}{c|}{\textcolor{orange}{\textbf{13}}}&\multicolumn{2}{c|}{All(1-13)}&\multicolumn{1}{c|}{\textcolor{beige}{\textbf{1-13}}} & \multicolumn{1}{c|}{\textcolor{orange}{\textbf{14}}}&\multicolumn{2}{c|}{All(1-14)}&Avg.& Avg.  \\
        Metrics   & \multicolumn{1}{c|}{Ma-F1}  & \multicolumn{1}{c|}{Ma-F1} & \multicolumn{1}{c}{Mi-F1}& \multicolumn{1}{c|}{Ma-F1}  & \multicolumn{1}{c|}{Ma-F1} & \multicolumn{1}{c|}{Ma-F1}& \multicolumn{1}{c}{Mi-F1}& \multicolumn{1}{c|}{Ma-F1}  &
        \multicolumn{1}{c|}{Ma-F1} & \multicolumn{1}{c|}{Ma-F1}&\multicolumn{1}{c}{Mi-F1}& \multicolumn{1}{c|}{Ma-F1}  & \multicolumn{1}{c|}{Ma-F1} & \multicolumn{1}{c|}{Ma-F1}&\multicolumn{1}{c}{Mi-F1}& \multicolumn{1}{c|}{Ma-F1} &
       Mi-F1& Ma-F1
        \\
 \hline
         FT+FL  & \multicolumn{1}{c|}{0.00} & \multicolumn{1}{c|}{61.90} & 5.34 & 5.63 & \multicolumn{1}{c|}{0.76} & \multicolumn{1}{c|}{35.04} & 16.65 & 3.62 & \multicolumn{1}{c|}{0.00} & \multicolumn{1}{c|}{46.96} & 5.68 & 3.61 & \multicolumn{1}{c|}{3.06} & \multicolumn{1}{c|}{48.78} & 21.60 & 6.33 & -- & -- \\

         PODNet+FL & \multicolumn{1}{c|}{32.84} & \multicolumn{1}{c|}{8.77} & 40.04 & 30.65 & \multicolumn{1}{c|}{6.88} & \multicolumn{1}{c|}{10.47} & 10.66 & 7.18 & \multicolumn{1}{c|}{6.56} & \multicolumn{1}{c|}{14.16} & 11.98 & 7.14 & \multicolumn{1}{c|}{5.17} & \multicolumn{1}{c|}{9.95} & 9.39 & 5.51 & -- & -- \\

     LUCIR+FL & \multicolumn{1}{c|}{\textcolor{blue}{\textbf{49.05}}} & \multicolumn{1}{c|}{63.25} & \textcolor{blue}{\textbf{70.67}} & 50.34 & \multicolumn{1}{c|}{43.02} & \multicolumn{1}{c|}{32.79} & 51.72 & 42.17 & \multicolumn{1}{c|}{35.89} & \multicolumn{1}{c|}{30.40} & 49.31 & 35.47 & \multicolumn{1}{c|}{28.17} & \multicolumn{1}{c|}{39.03} & 42.93 & 28.94 & -- & -- \\

      ST+FL & \multicolumn{1}{c|}{42.23} & \multicolumn{1}{c|}{63.04} & 64.50 & 44.12 & \multicolumn{1}{c|}{36.63} & \multicolumn{1}{c|}{35.96} & 52.49 & 36.57 & \multicolumn{1}{c|}{35.90} & \multicolumn{1}{c|}{42.11} & 51.58 & 36.38 & \multicolumn{1}{c|}{33.96} & \multicolumn{1}{c|}{51.26} & 50.63 & 35.19 & -- & -- \\

         ExtendNER+FL  & \multicolumn{1}{c|}{39.93} & \multicolumn{1}{c|}{63.42} & 64.18 & 42.07 & \multicolumn{1}{c|}{33.23} & \multicolumn{1}{c|}{35.06} & 51.41 & 33.38 & \multicolumn{1}{c|}{31.75} & \multicolumn{1}{c|}{43.02} & 50.43 & 32.62 & \multicolumn{1}{c|}{32.32} & \multicolumn{1}{c|}{50.31} & 49.35 & 33.60 & -- & -- \\

   CFNER+FL & \multicolumn{1}{c|}{47.07} & \multicolumn{1}{c|}{64.40} & 65.84 & 48.64 & \multicolumn{1}{c|}{\textcolor{blue}{\textbf{45.33}}} & \multicolumn{1}{c|}{\textcolor{blue}{\textbf{39.36}}} & 56.83 & \textcolor{blue}{\textbf{44.84}} & \multicolumn{1}{c|}{\textcolor{blue}{\textbf{43.78}}} & \multicolumn{1}{c|}{\textcolor{blue}{\textbf{54.53}}} & \textcolor{blue}{\textbf{57.14}} & \textcolor{blue}{\textbf{44.61}} & \multicolumn{1}{c|}{\textcolor{blue}{\textbf{43.09}}} & \multicolumn{1}{c|}{55.68} & \textcolor{blue}{\textbf{56.54}} & 43.99 & -- & -- \\

      CPFD+FL & \multicolumn{1}{c|}{48.94} & \multicolumn{1}{c|}{\textcolor{blue}{\textbf{70.27}}} & 66.83 & \textcolor{blue}{\textbf{50.88}} & \multicolumn{1}{c|}{44.12} & \multicolumn{1}{c|}{\textcolor{deepred}{\textbf{40.77}}} & \textcolor{blue}{\textbf{57.93}} & 43.84 & \multicolumn{1}{c|}{42.96} & \multicolumn{1}{c|}{\textcolor{deepred}{\textbf{58.59}}} & 56.10 & 44.17 & \multicolumn{1}{c|}{42.98} & \multicolumn{1}{c|}{\textcolor{blue}{\textbf{57.67}}} & 56.49 & \textcolor{blue}{\textbf{44.03}} & -- & -- \\

 \hline

 \textbf{LGFD (Ours)}   & \multicolumn{1}{c|}{\textcolor{deepred}{\textbf{55.07}}} & \multicolumn{1}{c|}{\textcolor{deepred}{\textbf{78.60}}} & \textcolor{deepred}{\textbf{72.15}} & \textcolor{deepred}{\textbf{57.21}} & \multicolumn{1}{c|}{\textcolor{deepred}{\textbf{47.58}}} & \multicolumn{1}{c|}{38.26} & \textcolor{deepred}{\textbf{58.41}} & \textcolor{deepred}{\textbf{46.80}} & \multicolumn{1}{c|}{\textcolor{deepred}{\textbf{46.63}}} & \multicolumn{1}{c|}{46.37} & \textcolor{deepred}{\textbf{58.49}} & \textcolor{deepred}{\textbf{46.61}} & \multicolumn{1}{c|}{\textcolor{deepred}{\textbf{46.28}}} & \multicolumn{1}{c|}{\textcolor{deepred}{\textbf{58.22}}} & \textcolor{deepred}{\textbf{58.09}} & \textcolor{deepred}{\textbf{47.13}} & -- & -- \\

\hline\hline
Task ID   & \multicolumn{4}{c|}{t=6}&\multicolumn{4}{c|}{t=7} &\multicolumn{4}{c|}{t=8}&\multicolumn{4}{c|}{t=9}& & \\
		Entity Type ID &\multicolumn{1}{c|}{\textcolor{beige}{\textbf{1-14}}} & \multicolumn{1}{c|}{\textcolor{orange}{\textbf{15}}}&\multicolumn{2}{c|}{All(1-15)} & \multicolumn{1}{c|}{\textcolor{beige}{\textbf{1-15}}} & \multicolumn{1}{c|}{\textcolor{orange}{\textbf{16}}}&\multicolumn{2}{c|}{All(1-16)}& \multicolumn{1}{c|}{\textcolor{beige}{\textbf{1-16}}} & \multicolumn{1}{c|}{\textcolor{orange}{\textbf{17}}}&\multicolumn{2}{c|}{All(1-17)}&\multicolumn{1}{c|}{\textcolor{beige}{\textbf{1-17}}} & \multicolumn{1}{c|}{\textcolor{orange}{\textbf{18}}}&\multicolumn{2}{c|}{All(1-18)}&Avg.& Avg.  \\
        Metrics   & \multicolumn{1}{c|}{Ma-F1}  & \multicolumn{1}{c|}{Ma-F1} & \multicolumn{1}{c}{Mi-F1}& \multicolumn{1}{c|}{Ma-F1}  & \multicolumn{1}{c|}{Ma-F1} & \multicolumn{1}{c|}{Ma-F1}& \multicolumn{1}{c}{Mi-F1}& \multicolumn{1}{c|}{Ma-F1}  &
        \multicolumn{1}{c|}{Ma-F1} & \multicolumn{1}{c|}{Ma-F1}&\multicolumn{1}{c}{Mi-F1}& \multicolumn{1}{c|}{Ma-F1}  & \multicolumn{1}{c|}{Ma-F1} & \multicolumn{1}{c|}{Ma-F1}&\multicolumn{1}{c}{Mi-F1}& \multicolumn{1}{c|}{Ma-F1} &
       Mi-F1& Ma-F1
        \\
    \hline
         FT+FL & \multicolumn{1}{c|}{0.00} & \multicolumn{1}{c|}{12.74} & 0.76 & 0.85 & \multicolumn{1}{c|}{0.00} & \multicolumn{1}{c|}{13.90} & 1.25 & 0.87 & \multicolumn{1}{c|}{0.00} & \multicolumn{1}{c|}{18.33} & 1.98 & 1.08 & \multicolumn{1}{c|}{0.00} & \multicolumn{1}{c|}{\textcolor{blue}{\textbf{7.32}}} & 0.73 & 0.41 & 6.75 & 2.80 \\

         PODNet+FL  & \multicolumn{1}{c|}{4.82} & \multicolumn{1}{c|}{0.58} & 3.13 & 4.54 & \multicolumn{1}{c|}{4.26} & \multicolumn{1}{c|}{1.61} & 1.45 & 4.10 & \multicolumn{1}{c|}{0.74} & \multicolumn{1}{c|}{0.75} & 1.08 & 0.74 & \multicolumn{1}{c|}{0.46} & \multicolumn{1}{c|}{0.23} & 0.57 & 0.45 & 9.79 & 7.54 \\

     LUCIR+FL & \multicolumn{1}{c|}{25.76} & \multicolumn{1}{c|}{5.28} & 40.29 & 24.39 & \multicolumn{1}{c|}{21.74} & \multicolumn{1}{c|}{5.51} & 37.05 & 20.73 & \multicolumn{1}{c|}{20.27} & \multicolumn{1}{c|}{6.83} & 32.98 & 19.48 & \multicolumn{1}{c|}{17.19} & \multicolumn{1}{c|}{3.99} & 26.31 & 16.46 & 43.91 & 29.75 \\

      ST+FL & \multicolumn{1}{c|}{36.09} & \multicolumn{1}{c|}{8.09} & 50.33 & 34.22 & \multicolumn{1}{c|}{33.06} & \multicolumn{1}{c|}{11.68} & 48.74 & 31.72 & \multicolumn{1}{c|}{31.98} & \multicolumn{1}{c|}{19.67} & 48.41 & 31.25 & \multicolumn{1}{c|}{31.26} & \multicolumn{1}{c|}{5.46} & 45.73 & 29.83 & 51.55 & 34.91 \\

         ExtendNER+FL & \multicolumn{1}{c|}{33.77} & \multicolumn{1}{c|}{9.23} & 48.41 & 32.13 & \multicolumn{1}{c|}{31.86} & \multicolumn{1}{c|}{10.84} & 47.16 & 30.54 & \multicolumn{1}{c|}{30.74} & \multicolumn{1}{c|}{19.62} & 46.35 & 30.09 & \multicolumn{1}{c|}{30.26} & \multicolumn{1}{c|}{5.04} & 43.59 & 28.86 & 50.11 & 32.91 \\

   CFNER+FL  & \multicolumn{1}{c|}{43.70} & \multicolumn{1}{c|}{9.41} & 54.74 & 41.41 & \multicolumn{1}{c|}{41.63} & \multicolumn{1}{c|}{\textcolor{blue}{\textbf{18.89}}} & 53.98 & 40.21 & \multicolumn{1}{c|}{39.53} & \multicolumn{1}{c|}{29.32} & 53.77 & 38.93 & \multicolumn{1}{c|}{38.24} & \multicolumn{1}{c|}{6.85} & 51.59 & 36.50 & 56.30 & 42.39 \\

      CPFD+FL & \multicolumn{1}{c|}{\textcolor{blue}{\textbf{44.93}}} & \multicolumn{1}{c|}{\textcolor{blue}{\textbf{17.61}}} & \textcolor{blue}{\textbf{56.38}} & \textcolor{blue}{\textbf{43.10}} & \multicolumn{1}{c|}{\textcolor{blue}{\textbf{43.46}}} & \multicolumn{1}{c|}{\textcolor{deepred}{\textbf{21.43}}} & \textcolor{blue}{\textbf{55.59}} & \textcolor{blue}{\textbf{42.08}} & \multicolumn{1}{c|}{\textcolor{blue}{\textbf{42.70}}} & \multicolumn{1}{c|}{\textcolor{blue}{\textbf{32.32}}} & \textcolor{blue}{\textbf{55.68}} & \textcolor{blue}{\textbf{42.09}} & \multicolumn{1}{c|}{\textcolor{blue}{\textbf{41.29}}} & \multicolumn{1}{c|}{\textcolor{deepred}{\textbf{8.08}}} & \textcolor{blue}{\textbf{54.20}} & \textcolor{blue}{\textbf{39.44}} & \textcolor{blue}{\textbf{57.40}} & \textcolor{blue}{\textbf{43.71}} \\

 \hline

 \textbf{LGFD (Ours)} & \multicolumn{1}{c|}{\textcolor{deepred}{\textbf{46.41}}} & \multicolumn{1}{c|}{\textcolor{deepred}{\textbf{30.77}}} & \textcolor{deepred}{\textbf{58.24}} & \textcolor{deepred}{\textbf{45.37}} & \multicolumn{1}{c|}{\textcolor{deepred}{\textbf{45.34}}} & \multicolumn{1}{c|}{15.11} & \textcolor{deepred}{\textbf{57.71}} & \textcolor{deepred}{\textbf{43.45}} & \multicolumn{1}{c|}{\textcolor{deepred}{\textbf{42.92}}} & \multicolumn{1}{c|}{\textcolor{deepred}{\textbf{32.90}}} & \textcolor{deepred}{\textbf{57.29}} & \textcolor{deepred}{\textbf{42.33}} & \multicolumn{1}{c|}{\textcolor{deepred}{\textbf{42.46}}} & \multicolumn{1}{c|}{7.00} & \textcolor{deepred}{\textbf{56.42}} & \textcolor{deepred}{\textbf{40.49}} & $\textbf{\textcolor{deepred}{59.60}}^{\dagger}$ & $\textbf{\textcolor{deepred}{46.17}}^{\dagger}$ \\

 \hline
    \textbf{Upper Bound}  &\multicolumn{1}{c|}{--}& \multicolumn{1}{c|}{--}&--&--& \multicolumn{1}{c|}{--}& \multicolumn{1}{c|}{--}&--&--&\multicolumn{1}{c|}{--}&\multicolumn{1}{c|}{--}&--&--&\multicolumn{1}{c|}{--}&\multicolumn{1}{c|}{--}&86.31&75.28&--&-- \\

\bottomrule

\end{tabular}
}
\label{tab:onto_10-1}
\end{table*}

\begin{table*}[t]
\centering
\caption{Comparison results in terms of Mi-F1 and Ma-F1 (\%) on the OntoNotes5 dataset under the setting of $10$-$2$. The \textcolor{beige}{\textbf{beige}} denotes the old entity types that have already been learned, and the \textcolor{orange}{\textbf{orange}} denotes the new entity types that are currently being learned. The \textcolor{deepred}{\textbf{red}} represents the highest result, and the \textcolor{blue}{\textbf{blue}} represents the second highest result. The markers $\natural$ and $\dagger$ refer to significant tests comparing with ST+FL and CPFD+FL, respectively ($p$-$value<0.05$).}
\resizebox{1.0\linewidth}{!}{
\begin{tabular}{c|cccc|cccc|cccc|cccc|>{\columncolor{lightgray}}c>{\columncolor{lightgray}}c}
		\toprule
		Task ID   & \multicolumn{4}{c|}{t=2}&\multicolumn{4}{c|}{t=3} &\multicolumn{4}{c|}{t=4}&\multicolumn{4}{c|}{t=5}& & \\
		Entity Type ID &\multicolumn{1}{c|}{\textcolor{beige}{\textbf{1-10}}} & \multicolumn{1}{c|}{\textcolor{orange}{\textbf{11-12}}}&\multicolumn{2}{c|}{All(1-12)} & \multicolumn{1}{c|}{\textcolor{beige}{\textbf{1-12}}} & \multicolumn{1}{c|}{\textcolor{orange}{\textbf{13-14}}}&\multicolumn{2}{c|}{All(1-14)}& \multicolumn{1}{c|}{\textcolor{beige}{\textbf{1-14 }}} & \multicolumn{1}{c|}{\textcolor{orange}{\textbf{15-16}}}&\multicolumn{2}{c|}{All(1-16)}&\multicolumn{1}{c|}{\textcolor{beige}{\textbf{1-16}}} & \multicolumn{1}{c|}{\textcolor{orange}{\textbf{17-18}}}&\multicolumn{2}{c|}{All(1-18)}&Avg.& Avg.  \\
        Metrics   & \multicolumn{1}{c|}{Ma-F1}  & \multicolumn{1}{c|}{Ma-F1} & \multicolumn{1}{c}{Mi-F1}& \multicolumn{1}{c|}{Ma-F1}  & \multicolumn{1}{c|}{Ma-F1} & \multicolumn{1}{c|}{Ma-F1}& \multicolumn{1}{c}{Mi-F1}& \multicolumn{1}{c|}{Ma-F1}  &
        \multicolumn{1}{c|}{Ma-F1} & \multicolumn{1}{c|}{Ma-F1}&\multicolumn{1}{c}{Mi-F1}& \multicolumn{1}{c|}{Ma-F1}  & \multicolumn{1}{c|}{Ma-F1} & \multicolumn{1}{c|}{Ma-F1}&\multicolumn{1}{c}{Mi-F1}& \multicolumn{1}{c|}{Ma-F1} &
       Mi-F1& Ma-F1
        \\
 \hline
         FT+FL   & \multicolumn{1}{c|}{0.00} & \multicolumn{1}{c|}{57.51} & 24.46 & 9.59 & \multicolumn{1}{c|}{0.00} & \multicolumn{1}{c|}{52.94} & 24.51 & 7.56 & \multicolumn{1}{c|}{0.00} & \multicolumn{1}{c|}{23.28} & 1.87 & 2.91 & \multicolumn{1}{c|}{0.11} & \multicolumn{1}{c|}{19.03} & 2.61 & 2.21 & 13.36 & 5.57 \\

         PODNet+FL   & \multicolumn{1}{c|}{31.19} & \multicolumn{1}{c|}{24.61} & 39.17 & 30.09 & \multicolumn{1}{c|}{20.14} & \multicolumn{1}{c|}{25.58} & 31.13 & 20.92 & \multicolumn{1}{c|}{11.75} & \multicolumn{1}{c|}{2.23} & 11.66 & 10.57 & \multicolumn{1}{c|}{10.30} & \multicolumn{1}{c|}{2.13} & 6.83 & 9.39 & 22.20 & 17.74\\

     LUCIR+FL  & \multicolumn{1}{c|}{50.98} & \multicolumn{1}{c|}{\textcolor{blue}{\textbf{66.58}}} & \textcolor{deepred}{\textbf{72.11}} & 53.58 & \multicolumn{1}{c|}{\textcolor{deepred}{\textbf{47.12}}} & \multicolumn{1}{c|}{54.39} & 64.22 & \textcolor{deepred}{\textbf{48.15}} & \multicolumn{1}{c|}{41.30} & \multicolumn{1}{c|}{15.97} & 55.10 & 38.14 & \multicolumn{1}{c|}{32.23} & \multicolumn{1}{c|}{12.93} & 48.95 & 30.09 & 60.10 & 42.49\\

      ST+FL & \multicolumn{1}{c|}{45.08} & \multicolumn{1}{c|}{66.07} & \textcolor{blue}{\textbf{71.66}} & 48.58 & \multicolumn{1}{c|}{43.90} & \multicolumn{1}{c|}{\textcolor{deepred}{\textbf{59.03}}} & \textcolor{deepred}{\textbf{66.00}} & 46.06 & \multicolumn{1}{c|}{39.97} & \multicolumn{1}{c|}{20.42} & 54.38 & 37.52 & \multicolumn{1}{c|}{37.58} & \multicolumn{1}{c|}{20.98} & 52.67 & 35.74 & \textcolor{blue}{\textbf{61.17}} & 41.98\\

     ExtendNER+FL  & \multicolumn{1}{c|}{46.48} & \multicolumn{1}{c|}{66.00} & 71.31 & 49.74 & \multicolumn{1}{c|}{43.61} & \multicolumn{1}{c|}{\textcolor{blue}{\textbf{57.34}}} & \textcolor{blue}{\textbf{64.73}} & 45.57 & \multicolumn{1}{c|}{39.25} & \multicolumn{1}{c|}{21.35} & 54.24 & 37.02 & \multicolumn{1}{c|}{37.47} & \multicolumn{1}{c|}{23.64} & 52.64 & 35.94 & 60.73 & 42.06\\

   CFNER+FL  & \multicolumn{1}{c|}{50.17} & \multicolumn{1}{c|}{63.59} & 67.03 & 52.41 & \multicolumn{1}{c|}{44.70} & \multicolumn{1}{c|}{52.20} & 57.92 & 45.78 & \multicolumn{1}{c|}{45.99} & \multicolumn{1}{c|}{27.18} & 57.41 & 43.64 & \multicolumn{1}{c|}{43.58} & \multicolumn{1}{c|}{\textcolor{blue}{\textbf{24.49}}} & 56.27 & 41.46 & 59.66 & 45.82\\

      CPFD+FL  & \multicolumn{1}{c|}{\textcolor{blue}{\textbf{52.54}}} & \multicolumn{1}{c|}{65.48} & 68.46 & \textcolor{blue}{\textbf{54.70}} & \multicolumn{1}{c|}{45.80} & \multicolumn{1}{c|}{52.33} & 58.73 & 46.73 & \multicolumn{1}{c|}{\textcolor{deepred}{\textbf{47.41}}} & \multicolumn{1}{c|}{\textcolor{blue}{\textbf{31.22}}} & \textcolor{blue}{\textbf{58.63}} & \textcolor{blue}{\textbf{45.39}} & \multicolumn{1}{c|}{\textcolor{blue}{\textbf{44.59}}} & \multicolumn{1}{c|}{\textcolor{deepred}{\textbf{25.02}}} & \textcolor{blue}{\textbf{57.60}} & \textcolor{blue}{\textbf{42.41}} & 60.85 & \textcolor{blue}{\textbf{47.31}}\\
 \hline

 \textbf{LGFD (Ours)} & \multicolumn{1}{c|}{\textcolor{deepred}{\textbf{56.02}}} & \multicolumn{1}{c|}{\textcolor{deepred}{\textbf{69.38}}} & 71.35 & \textcolor{deepred}{\textbf{58.24}} & \multicolumn{1}{c|}{\textcolor{blue}{\textbf{46.71}}} & \multicolumn{1}{c|}{50.13} & 60.08 & \textcolor{blue}{\textbf{47.20}} & \multicolumn{1}{c|}{\textcolor{blue}{\textbf{47.15}}} & \multicolumn{1}{c|}{\textcolor{deepred}{\textbf{34.04}}} & \textcolor{deepred}{\textbf{59.55}} & \textcolor{deepred}{\textbf{45.51}} & \multicolumn{1}{c|}{\textcolor{deepred}{\textbf{44.78}}} & \multicolumn{1}{c|}{24.37} & \textcolor{deepred}{\textbf{58.78}} & \textcolor{deepred}{\textbf{42.51}} & $\textbf{\textcolor{deepred}{62.44}}^{\natural}$ & $\textbf{\textcolor{deepred}{48.37}}^{\dagger}$\\

     \hline
    \textbf{Upper Bound}  &\multicolumn{1}{c|}{--}& \multicolumn{1}{c|}{--}&--&--& \multicolumn{1}{c|}{--}& \multicolumn{1}{c|}{--}&--&--&\multicolumn{1}{c|}{--}&\multicolumn{1}{c|}{--}&--&--&\multicolumn{1}{c|}{--}&\multicolumn{1}{c|}{--}&86.31&75.28&--&-- \\
	
		\bottomrule
\end{tabular}
}
\label{tab:onto_10-2}
\end{table*}

\subsection{Baselines} 

We consider the following baselines, which includes SOTA INER methods: ExtendNER \cite{monaikul2021continual}, CFNER \cite{zheng2022distilling}, and CPFD \cite{zhang2023continual}. Moreover, we include the lower bound method, Fine-Tuning (FT), which directly employs new data for fine-tuning the model without utilizing any anti-forgetting techniques. Furthermore, we incorporate incremental learning methods from the computer vision field, such as PODNet \cite{DBLP:conf/eccv/DouillardCORV20}, LUCIR \cite{DBLP:conf/cvpr/HouPLWL19}, and Self-Training (ST) \cite{DBLP:journals/corr/abs-1909-08383}, which are adapted to the INER scenario. The detailed introductions to these baselines are as follows:

\paragraph{PODNet} \cite{DBLP:conf/eccv/DouillardCORV20} addresses the challenge of catastrophic forgetting in incremantal learning for image classification, which is adapted to the scenario of INER. The model's overall loss includes both classification and distillation losses. For classification, PODNet employs the neighborhood component analysis loss instead of the conventional cross-entropy loss. In distillation loss computation, PODNet imposes constraints on the output of each intermediate layer.

\paragraph{LUCIR} \cite{DBLP:conf/cvpr/HouPLWL19} establishes a framework for incremental learning in image classification tasks, transferred to the INER scenario. Similar to PODNet, it addresses catastrophic forgetting. Its overall loss comprises three components: (1) cross-entropy loss on samples with new entity types; (2) distillation loss between features extracted by the old and new models; and (3) margin-ranking loss on reserved samples for old types.

\paragraph{ST} \cite{DBLP:conf/wacv/RosenbergHS05,DBLP:journals/corr/abs-1909-08383} directly uses the pre-existing model to label the non-entity type tokens with their respective old entity types. The new model then undergoes training on novel data, incorporating annotations for all entity types. The objective is to minimize cross-entropy loss across all entity types, ensuring comprehensive and effective model training.

\paragraph{ExtendNER} \cite{monaikul2021continual} applies KD to INER, akin to ST, but computes cross-entropy loss for entity type tokens and KL divergence loss for the non-entity type tokens. During training, ExtendNER minimizes the sum of cross-entropy loss and KL divergence loss.

\paragraph{CFNER} \cite{zheng2022distilling}, based on ExtendNER, introduces a causal framework, extracting causal effects from the non-entity type tokens. It initially utilizes the old model to recognize non-entity type tokens belonging to previous entity types and employs curriculum learning to mitigate recognition errors.

\paragraph{CPFD} \cite{zhang2023continual} addresses catastrophic forgetting and the semantic shift problem of the non-entity type tokens in INER. It introduces a pooled feature distillation loss to balance stability and plasticity and proposes a confidence-based pseudo-labeling strategy to reduce label noise and address the semantic shift problem.

For fair comparisons with these baseline INER methods under the FINER setup, we utilize the same backbone encoder (\emph{i.e.}, bert-base-cased \cite{kenton2019bert}) and ``BIO'' labeling schema for all datasets. This schema assigns two labels to each entity type: B-entity (for the beginning of an entity) and I-entity (for the inside of an entity).

\subsection{Implementation Details} 

Our model is implemented in the PyTorch framework~\cite{paszke2019pytorch} on top of the BERT Huggingface implementation~\cite{wolf2019huggingface}, with the default hyper-parameters: hidden dimensions $d_h$ are $768$, and the max sequence length is $512$.
We utilize the SGD optimizer with an initial learning rate of $2\times10^{-3}$ to train the first base task and $4\times10^{-4}$ for subsequent incremental tasks.
The batch size, global round for each task $R$, feature groups $G$ and trade-off hyper-parameters $\lambda_1$ and $\lambda_2$ are set to $16$, $5$, $12$, $2$, and $0.02$, respectively. 
Please note that we do not meticulously search for optimal hyper-parameters, and instead rely on default values across all experiments. Consequently, finely tuning the hyper-parameters could potentially enhance performance notably on particular datasets and configurations.
All experiments are conducted on an $40$GB NVIDIA A100 GPU. Each experiment is run $3$ times with $3$ different random seeds ($2023$, $2024$, $2025$), and the averaged results are reported to ensure statistical robustness.

\begin{table*}[t]
	\centering
   \caption{The ablation study of our LGFD model under the 8-1 and 8-2 FINER settings on the I2B2 dataset, {as well as the 10-1 and 10-2 FINER settings on the OntoNotes5 dataset.} Compared with our LGFD model, all ablation variants significantly degrade FINER performance, verifying the importance of all components to address FINER collaboratively. The \textbf{Bold} denotes the best performance.}
	\resizebox{1.0\linewidth}{!}{
			\begin{tabular}{c|cc|cc|cc|cc|cc}
				\toprule
				& \multicolumn{2}{c|}{Variants} &\multicolumn{2}{c|}{$8$-$1$} & \multicolumn{2}{c|}{$8$-$2$} &\multicolumn{2}{c|}{{$10$-$1$}}&\multicolumn{2}{c}{{$10$-$2$}} \\
				
				 & SKD& ITC  & Avg. Mi-F1 & Avg. Ma-F1 & Avg. Mi-F1 & Avg.Ma-F1& Avg. Mi-F1 & Avg. Ma-F1& Avg. Mi-F1 & Avg. Ma-F1 \\
				\hline
    	 LGFD w/o $\mathcal{L}_{\text{SKD}}$ & \XSolidBrush & \CheckmarkBold  &37.29  & 25.76 & 44.02 & 28.26 &56.26  & 44.14 & 59.63 & 46.17\\

       LGFD w/ $\mathcal{L}_{\text{FD}}$ & \XSolidBrush & \CheckmarkBold  & 40.11 & 27.40 & 46.35 & 30.98 & 58.87 & 45.48 & 61.52 & 47.91\\
       LGFD w/o  $\mathcal{L}_{\text{ITC}}$ & \CheckmarkBold & \XSolidBrush  & {37.83} & {26.86} & {45.28} & {30.14} & {58.41} & {45.19} & {61.29} & {47.79}\\
				
\rowcolor{lightgray} \textbf{LGFD (Ours)} &  \CheckmarkBold & \CheckmarkBold & \textbf{43.79} & \textbf{29.79} & \textbf{48.08} & \textbf{33.05} & \textbf{59.60} & \textbf{46.17} & \textbf{62.44}& \textbf{48.37}  \\
				\bottomrule
		\end{tabular}}
  	
		\label{tab:ablation_studies}
	\end{table*}

\section{Experimental Results}

\subsection{Main Results}

We conduct extensive experiments on the I2B2 and OnteNotes5 datasets to analyze the superiority of our LGFD model under various FINER settings. 
The experimental results are presented in Tables~\ref{tab:i2b2_8-1},~\ref{tab:i2b2_8-2},~\ref{tab:onto_10-1}, and~\ref{tab:onto_10-2}. 
Our LGFD model significantly outperforms previous SOTA INER methods in the final average Mi-F1 and Ma-F1 scores across various FINER settings, with the majority of intermediate results also leading, demonstrating the effectiveness of our LGFD model in learning a global INER model via collaborative training of local models while preserving privacy. 

Specifically, in Table~\ref{tab:i2b2_8-1}, our LGFD model shows significant enhancements over the SOTA baseline CPFD+FL, with improvements of 8.99\% and 3.49\% in final average Mi-F1 and Ma-F1 scores, respectively, under the $8$-$1$ FINER settings of the I2B2 dataset.
Table~\ref{tab:i2b2_8-2} illustrates similar improvements, where our LGFD model outperforms the SOTA baseline CPFD+FL by 3.27\% in final average Mi-F1 score and the SOTA baseline CFNER+FL by 2.78\% in final average Ma-F1 score under the $8$-$2$ FINER settings of the I2B2 dataset.
Furthermore, in Table~\ref{tab:onto_10-1}, our LGFD model exhibits enhancements over the SOTA baseline CPFD+FL by 2.20\% in final average Mi-F1 score and 2.46\% in final average Ma-F1 score under the $10$-$1$ FINER settings of the OntoNotes5 dataset.
Lastly, as shown in Table~\ref{tab:onto_10-2}, our LGFD model displays improvements over the SOTA baseline ST+FL by 1.27\% in final average Mi-F1 score and over the SOTA baseline CPFD+FL by 1.06\% in final average Ma-F1 score under the $10$-$2$ FINER settings of the OntoNotes5 dataset.

{Additionally, we establish an Upper Bound in Tables \ref{tab:i2b2_8-1}, \ref{tab:i2b2_8-2}, \ref{tab:onto_10-1}, and \ref{tab:onto_10-2} by performing a one-time supervised fine-tuning on the complete I2B2 and OntoNotes5 datasets, covering all entity types. This serves as a performance upper bound for the \textbf{final step} of FINER. The significant performance gap between the upper bound (\emph{e.g.}, common supervised fine-tuning) and our proposed LGFD method arises from the unique challenges inherent in the FINER setup, which closely mirrors real-world scenarios. Specifically, two major challenges contribute to this gap: (1) \textbf{Incremental Learning within Clients}: In FINER, clients learn from streaming data, which introduces severe catastrophic forgetting issues—challenges not present in standard supervised fine-tuning setups. (2) \textbf{Federated Learning across Clients}: Unlike centralized training, FINER requires federated learning among clients, where data is Non-IID. This poses substantial challenges in aggregating and synchronizing updates from different clients. These factors make FINER inherently more complex and challenging, resulting in the observed performance gap. Nonetheless, the proposed LGFD method is practically effective for real-world FINER scenarios, as it significantly outperforms previous SOTA approaches. While there is still room for improvement, we introduce this benchmark to inspire future research aimed at developing more advanced methods to narrow the gap with the upper bound.}

\subsection{{Ablation Study}}

To assess the effectiveness of each module in our LGFD model, Table~\ref{tab:ablation_studies} presents ablation experiments conducted under the $8$-$1$ and $8$-$2$ FINER settings of the I2B2 dataset, {as well as the $10$-$1$ and $10$-$2$ FINER settings on the OntoNotes5 dataset.} When compared with our LGFD, all ablation variants exhibit severe degradation in performance, highlighting the importance of all modules in addressing the FINER setup collaboratively. Additionally, when replacing $\mathcal{L}_{\mr{SKD}}$ with $\mathcal{L}_{\mr{FD}}$, a decrease in performance is observed, indicating that SKD is more effective than FD in preserving previously learned knowledge. By retaining important features and reducing the interference of redundant and irrelevant features, SKD demonstrates its effectiveness.

\begin{table}[tbp]
  \centering
\caption{{The sensitivity analysis of hyper-parameter $\lambda_1$ on the I2B2 dataset \cite{murphy2010serving} under the $8$-$2$ setting.}}
  \resizebox{0.9\linewidth}{!}{
    \begin{tabular}{ccc}
    \toprule
    $\lambda_1$ in Equation (\ref{eq:overall_optimization}) & Avg. Mi-F1 & Avg. Ma-F1  \\
    \midrule

1.0 & 46.98 & 32.56\\

1.5 & 47.61 & 33.72 \\

2.0 & 48.08 & 33.05 \\

2.5 & 47.14 &  32.38\\
   
    \bottomrule
    \end{tabular}%
    } 
  \label{hyper}%
\end{table}%

\subsection{Sensitivity Analysis of the Hyper-parameters}

{We conduct a sensitivity analysis of hyper-parameter $\lambda_1$ in the $8$-$2$ setting on the I2B2 dataset. This hyper-parameter primarily serves to balance the loss terms in our model. As shown in Table \ref{hyper}, we observed that varying $\lambda_1$ within a certain range, particularly around a value of $2$, did not significantly impact model performance. This stability in performance across different values of $\lambda_1$ suggests that our model's results are robust and not highly sensitive to the specific choice of this hyper-parameter.}

\subsection{Stability Concerning Entity Type Orders}

\begin{figure*}[t]
\centering
  \includegraphics[width=0.88\linewidth]{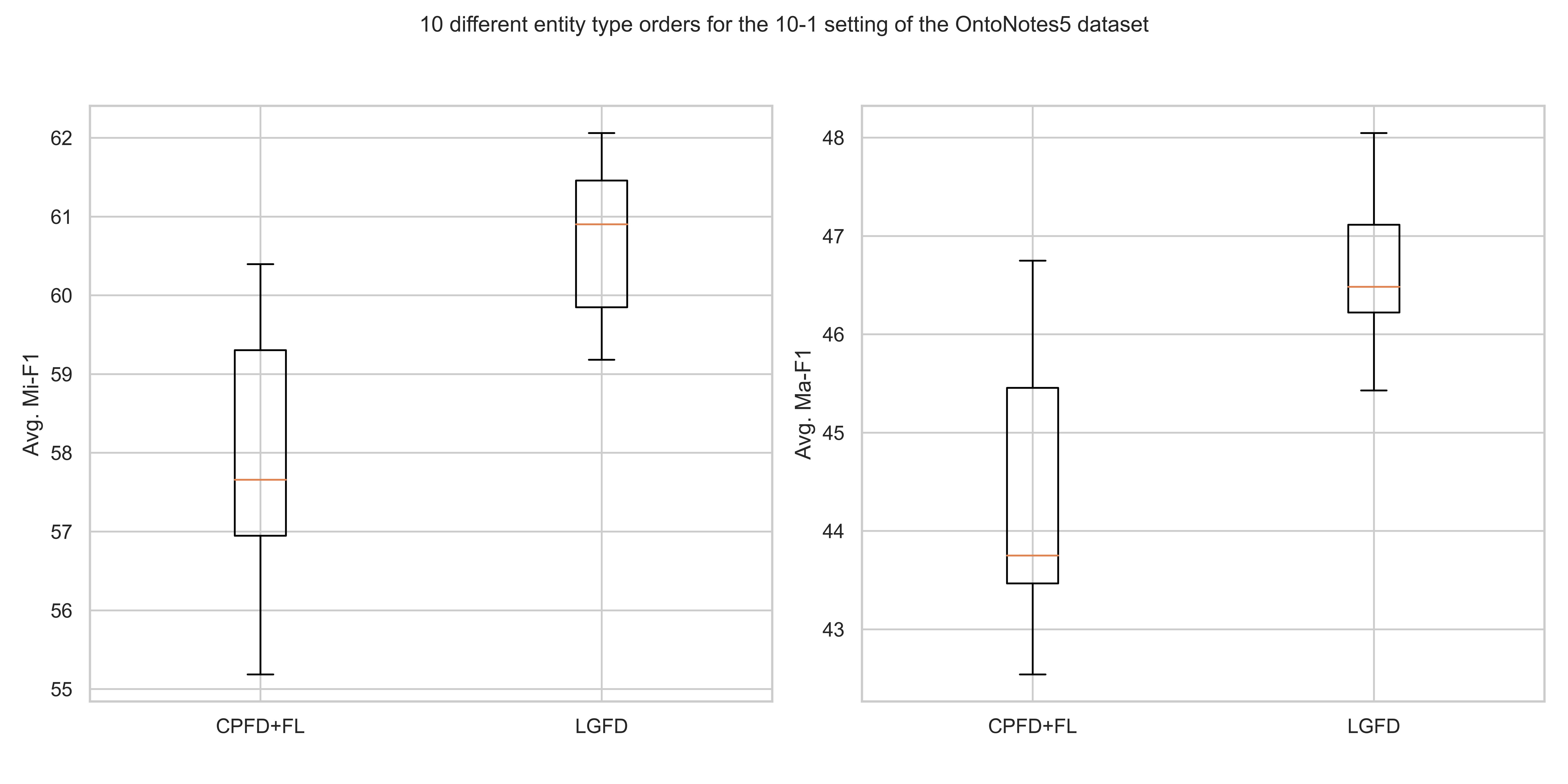} 
    \caption{The boxplots display the final average Mi-F1 and Ma-F1 scores across all tasks for $10$ random entity type orders. LGFD is significantly better and more stable than CPFD+FL.}
\label{fig:order}
\end{figure*}

{In real-world applications, the optimal order of entity types is unknown in advance. Consequently, an effective FINER method should ideally be as insensitive as possible to variations in entity type order. In our previous experiments, the entity type order was kept fixed (specifically, alphabetical order). To assess the robustness of our model under different orderings, we conducted experiments with $10$ random permutations of entity type order in the $10$-$1$ setting of the OntoNotes5 dataset, with results displayed in Figure \ref{fig:order} as boxplots. This figure presents the final average Mi-F1 and Ma-F1 scores across all tasks (excluding the first). As illustrated in Figure \ref{fig:order}, the boxplot for LGFD exhibits higher values, a more compact data distribution, and a narrower range of variation compared to CPFD+FL. These findings demonstrate that our proposed LGFD model not only outperforms but is also significantly more stable than the previous SOTA approach, CPFD+FL.}

\section{Conclusion}

In this paper, we introduce a FINER setup and develop a novel LGFD model to tackle intra-client and inter-client heterogeneous forgetting of old entity types. To address intra-client forgetting, we devise a SKD loss and a pseudo-label-guided ITC loss, effectively retaining previously learned knowledge within local client. Moreover, we propose a task switching monitor to handle inter-client forgetting, which automatically identifies new entity types and stores the latest old global model for knowledge distillation and pseudo-labeling. Comparative results demonstrate the superiority of our LGFD model in addressing the FINER setup.
In the future, we will consider using only few samples of new entity types to address intra-client and inter-client heterogeneous forgetting.

\bibliographystyle{IEEEtran}
\bibliography{reference}

\begin{IEEEbiography}[{\includegraphics[width=1in,height=1.25in,clip,keepaspectratio]{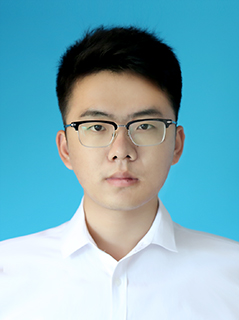}}]{Duzhen Zhang} received his B.Sc. degree from Shandong University in June 2019. He completed his Ph.D. at the Institute of Automation, Chinese Academy of Sciences in June 2024. Since September 2024, he has been a postdoctoral researcher at the Mohamed bin Zayed University of Artificial Intelligence. His current research interests include large language models, continual learning, multi-modal learning, and AI for science.
\end{IEEEbiography} 

\begin{IEEEbiography}
[{\includegraphics[width=1in,height=1.25in,clip,keepaspectratio]{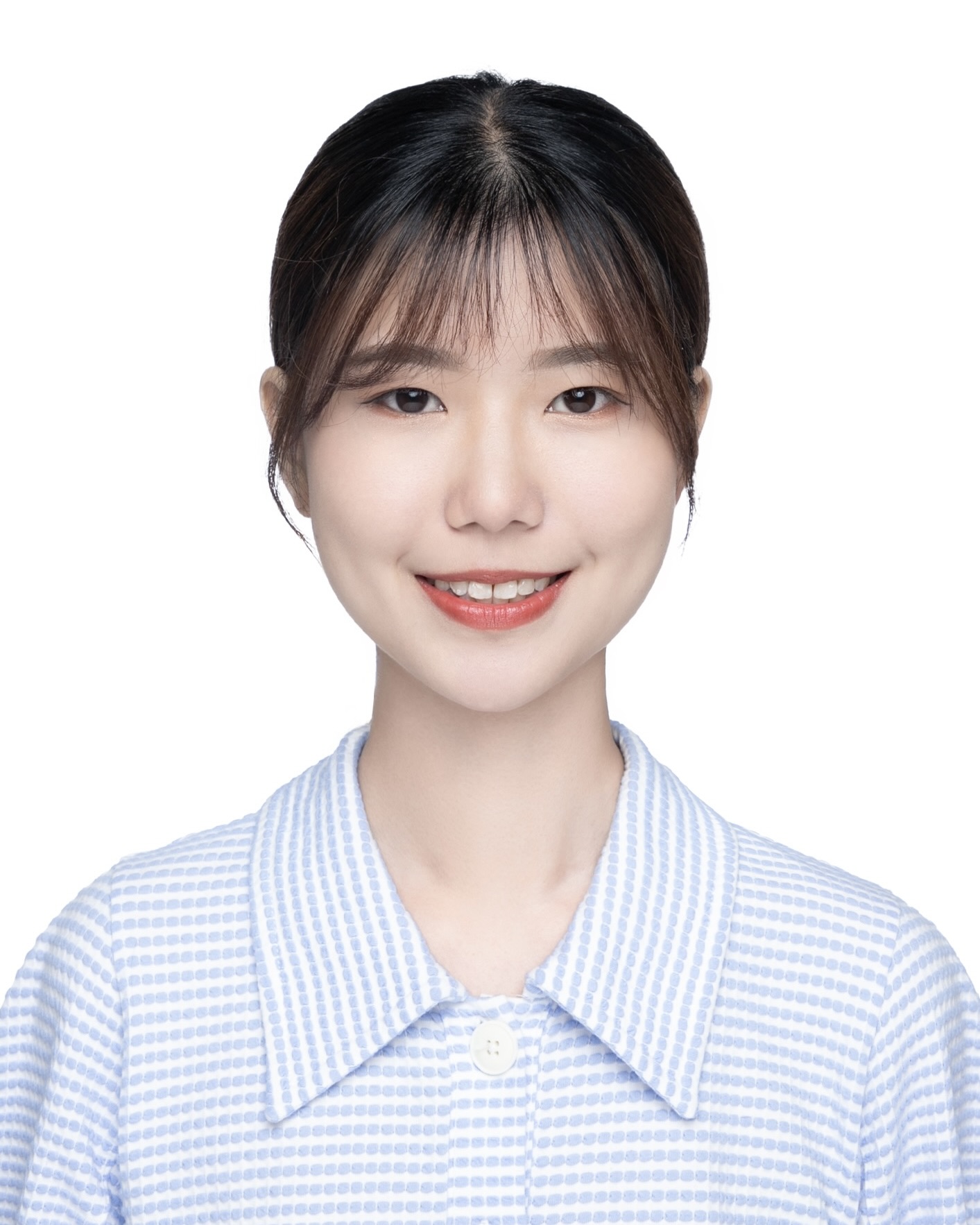}}]{Yahan Yu} received her B.S. degree from Nanjing University of Aeronautics and Astronautics in 2020 and her M.S. degree in Control Science from University of Chinese Academy of Sciences in 2023.
She is currently a Ph.D. student at Kyoto University. Her research interests include natural language processing and continual learning.
\end{IEEEbiography} 

\begin{IEEEbiography}[{\includegraphics[width=1in,height=1.25in,clip,keepaspectratio]{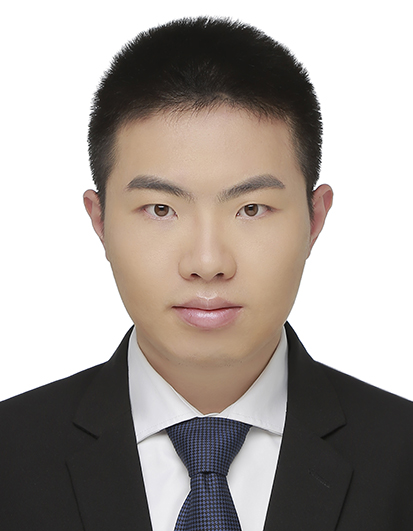}}]{Chenxing Li} received his B.Sc. degree at the North China Electric Power University, China, in 2015. He completed his Ph.D. at the Institute of Automation, Chinese Academy of Sciences in 2020. His current research interests include far-field speech recognition, speech enhancement, and speech separation.
\end{IEEEbiography} 

\begin{IEEEbiography}[{\includegraphics[width=1in,height=1.25in,clip,keepaspectratio]{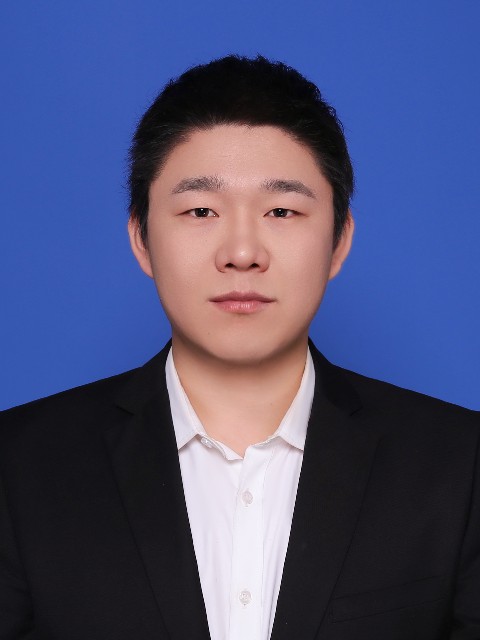}}]{Jiahua Dong} is a postdoctoral researcher in the Mohamed bin Zayed University of Artificial Intelligence, Abu Dhabi, United Arab Emirates. He received the Ph.D. degree from the Shenyang Institute of Automation, Chinese Academy of Sciences in 2024. He visited the Computer Vision Lab, ETH Zurich, Switzerland from April 2022 to August 2022, and Max Planck Institute for Informatics, Germany from September 2022 to January 2023. Before that, he received the B.S. degree from Jilin University in 2017. His current research interests include computer vision, machine learning and medical image analysis.
\end{IEEEbiography}

\begin{IEEEbiography}[{\includegraphics[width=1in,height=1.25in,clip,keepaspectratio]{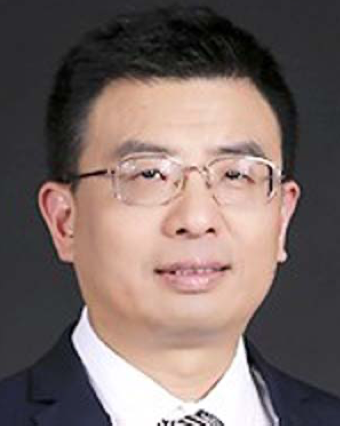}}]{Dong Yu} (Fellow, IEEE) with the Tencent AI Lab as a distinguished scientist and vice general Manager. Prior to joining Tencent in 2017, he was a Principal Researcher with Microsoft Research (Redmond), where he has been since 1998. He has authored or coauthored two monographs and more than 300 papers. His research interests include speech recognition and processing and natural language processing. His works have been widely cited and recognized by the prestigious IEEE Signal Processing Society best transaction paper award in 2013, 2016, 2020, and 2022, the 2021 NAACL best long paper award, 2022 IEEE Signal Processing Magazine best paper award, and 2022 IEEE Signal Processing Magazine best column award. Dr. Dong Yu was the Chair of the IEEE Speech and Language Processing Technical Committee during 2021–2022. He was on the editorial boards of numerous journals and magazines, as well as on the organizing and technical committees of various conferences and workshops. He is currently an ACM/IEEE/ISCA Fellow.
\end{IEEEbiography}

\end{document}